\documentclass{article}

\usepackage{PRIMEarxiv}

\usepackage[utf8]{inputenc} 
\usepackage[T1]{fontenc}    
\usepackage{hyperref}       
\usepackage{url}            
\usepackage{booktabs}       
\usepackage{amsfonts}       
\usepackage{nicefrac}       
\usepackage{microtype}      
\usepackage{lipsum}
\usepackage{fancyhdr}       
\usepackage{graphicx}       
\graphicspath{{media/}}     

\usepackage{natbib}
\usepackage{amsmath} 
\usepackage{amssymb}
\usepackage{xcolor}
\usepackage{subcaption}
\usepackage{tabularx}
\usepackage{multirow}
\usepackage{algorithm}
\usepackage{algpseudocode}
\usepackage{comment}
\usepackage{float}
\usepackage{booktabs}
\usepackage{bm}  
\usepackage{placeins}

\title{Gated X-TFC: Soft Domain Decomposition for Forward and Inverse Problems in Sharp-Gradient PDEs}

\author{
  Vikas Dwivedi\thanks{Corresponding Author} \\
  CREATIS Biomedical Imaging Laboratory \\
  INSA, CNRS UMR 5220, Inserm, Universit´e Lyon 1 \\
  Lyon 69621, France\\
  \texttt{vikas.dwivedi@creatis.insa-lyon.fr} \\
   \And
  Enrico Schiassi \\
  Department of Industrial Engineering \\
  University of Bologna \\
  Bologna 40126, Italy\\
  \texttt{enrico.schiassi2@unibo.it} \\
     \And
  Monica Sigovan \\
  CREATIS Biomedical Imaging Laboratory \\
  INSA, CNRS UMR 5220, Inserm, Universit´e Lyon 1 \\
  Lyon 69621, France\\
  \texttt{monica.sigovan@insa-lyon.fr} \\
    \And
  Bruno Sixou \\
  CREATIS Biomedical Imaging Laboratory \\
  INSA, CNRS UMR 5220, Inserm, Universit´e Lyon 1 \\
  Lyon 69621, France\\
  \texttt{bruno.sixou@insa-lyon.fr} \\
}

\begin{document}
\maketitle

\begin{abstract}
Physics-informed neural networks (PINNs) and related methods struggle to resolve sharp gradients in singularly perturbed boundary value problems without resorting to some form of domain decomposition, which often introduce complex interface penalties. While the Extreme Theory of Functional Connections (X–TFC) avoids multi-objective optimization by employing exact boundary condition enforcement, it remains computationally inefficient for boundary layers and incompatible with decomposition. We propose Gated X-TFC, a novel framework for both forward and inverse problems, that overcomes these limitations through a soft, learned domain decomposition. Our method replaces hard interfaces with a differentiable logistic gate that dynamically adapts radial basis function (RBF) kernel widths across the domain, eliminating the need for interface penalties. This approach yields not only superior accuracy but also dramatic improvements in computational efficiency: on a benchmark one dimensional (1D) convection-diffusion, Gated X-TFC achieves an order-of-magnitude lower error than standard X–TFC while using 80$\%$ fewer collocation points and reducing training time by 66 $\%$. In addition, we introduce an operator-conditioned meta-learning layer that learns a probabilistic mapping from PDE parameters to optimal gate configurations, enabling fast, uncertainty-aware warm-starting for new problem instances. We further demonstrate scalability to multiple subdomains and higher dimensions by solving a twin boundary–layer equation and a 2D Poisson problem with a sharp Gaussian source. Overall, Gated X--TFC delivers a simple alternative alternative to PINNs that is both accurate and computationally efficient for challenging boundary–layer regimes. Future work will focus on nonlinear problems. 
\end{abstract}


\section{Introduction}

Physics-informed neural networks (PINNs)~\cite{RAISSI2019686}, along with operator-learning techniques such as DeepONets~\cite{Lu2021} and Fourier Neural Operators (FNO)~\cite{li2021fourierneuraloperatorparametric}, have demonstrated great potential for solving forward and inverse problems involving partial differential equations (PDEs). A notable limitation, however, emerges when these methods are applied to singularly perturbed systems, which exhibit sharp spatial gradients or stiff temporal behavior, often leading to degraded performance~\cite{doi:10.1137/23M1623707,du2023approximationgeneralizationdeeponetslearning,you2024mscale,li2024componentfourierneuraloperator,lin2023operator}. This decline is largely attributed to spectral bias~\cite{pmlr-v97-rahaman19a,NEURIPS2020_55053683,CiCP-28-5}, which impedes the model's capacity to learn high-frequency, localized features essential for accurately representing such systems.

In response, a range of specialized PINN variants has been proposed. These include NTK-guided PINNs, which use neural tangent kernel theory to improve training dynamics and counter spectral bias~\cite{WANG2022110768}; self-adaptive PINNs that integrate trainable weights to focus attention on stiff regions~\cite{MCCLENNY2023111722}; and semi-analytic PINNs, which augment the solution with boundary-layer correctors to resolve singular behavior~\cite{Gie21092024}. Other approaches involve variable scaling to intensify focus on sensitive regions (VS-PINNs)~\cite{KO2025113860}, curriculum-based training that gradually introduces stiffness~\cite{NEURIPS2021_df438e52}, the use of pretrained operators to boost convergence (operator learning-enhanced PINNs)~\cite{lin2023operatorlearningenhancedphysicsinformed}, and theory-guided architectures that decouple boundary layers into separate networks~\cite{ARZANI2023111768}. While their strategies differ—spanning architectural changes, loss reweighting, sophisticated sampling, and hypothesis space refinement—these methods are unified in their goal to dynamically identify and adapt to stiffness. Nevertheless, these enhancements frequently come with substantial computational overhead and persistent issues regarding stability.

Recent research has sought to improve the speed and accuracy of scientific machine learning by moving away from computationally expensive gradient-based optimization. This has spurred interest in methods like Extreme Learning Machines (ELMs) \citep{HUANG2006489} and physics-informed ELMs (PIELM) \citep{DWIVEDI202096}, alongside the Extreme Theory of Functional Connections (X-TFC) \citep{SCHIASSI2021334}. In a related development, Bayesian approaches to PIELM \citep{LIU2023126425} have emerged. These methods recast forward and inverse problems into a linear-Gaussian framework, which enables evidence-based model selection and delivers calibrated uncertainty estimates, all without relying on deep backpropagation. 

Despite the speed advantage, these methods have their own limitations. For example, PIELMs are highly sensitive to hyperparameters in stiff regimes and struggle with inverse problems due to the inherent nonlinearity, even when the forward PDE is linear \citep{DONG2022111290}. Some recent works on PIELM with physics-aware hyperparameter selection include curriculum-learning \citep{DWIVEDI2025130924} and kernel adaptive PIELMs \citep{dwivedi2025kerneladaptivepielmsforwardinverse}. 

Among these methods, X-TFC stands out for its minimal objective: its constrained trial function satisfies boundary conditions exactly, eliminating boundary penalties altogether. However, even state-of-the-art (SOTA) X-TFC can be computationally inefficient for sharp gradients: capturing boundary-layer structure requires large number of neurons \citep{DEFLORIO2024115396} even in 1D, making scaling to higher dimensions problematic. A common workaround in PINNs and PIELMs is domain decomposition, but enforcing continuity and smoothness between subdomains typically introduces interface loss terms, complicating the objective and tuning. Moreover, extending X-TFC with a hard interface is nontrivial, since deriving constrained expressions tied to arbitrary internal boundaries is cumbersome.

We address these limitations with \emph{Gated X--TFC}, which performs a soft domain decomposition. The domain is adaptively split, and a logistic gate smoothly modulates RBF widths across the split. A single global X--TFC trial solution spans the interval and satisfies boundary conditions exactly, so no interface penalties are needed. The interface location and the gate’s transition scale are learned. Forward solves estimate coefficients by least squares on the strong–form residual; inverse problems stack data and PDE residual in a whitened linear model with evidence–based regularization and Bayesian optimization over the PDE and gate hyperparameters. Beyond the base solver, we introduce an operator–conditioned meta–learner that sweeps over PDE parameter offline and fits a probabilistic map between PDE parameter and the learned gating hyperparameters. At test time this provides uncertainty-aware warm starts and tight search windows, cutting outer-loop search cost while preserving accuracy.

\paragraph{Contributions}
\begin{itemize}
	\item \textbf{Gated X--TFC:} A soft, gate–based domain decomposition for X--TFC that enforces boundary conditions exactly and removes interface penalties while resolving thin layers.
	\item \textbf{Unified forward \& inverse solver:} Strong–form residual least squares for forward problems; a whitened linear–Gaussian formulation with evidence–based regularization and Bayesian optimization for inverse identification.
	\item \textbf{Operator–conditioned meta–learning:} A probabilistic regression from PDE parameters to gate hyperparameters that yields uncertainty–aware warm starts and \emph{tight} search bounds for outer optimization, leading to faster and more reliable convergence than global, hand–tuned boxes.
	\item \textbf{Empirical gains:} On the 1D convection–diffusion benchmark with small diffusion, Gated X--TFC attains accurate forward and inverse solutions and improves the speed/accuracy tradeoff compared to PINNs, PIELMs, Deep-TFC and X--TFC.
	\item \textbf{Practical and extensible:} The method naturally scales to multiple splits and higher dimensions, as illustrated on a twin boundary–layer equation and a 2D Poisson problem with a sharp Gaussian source.
\end{itemize}

The organization of this paper is as follows: Section~\ref{sec:X--TFC_Review} briefly reviews PINN, PIELM and X--TFC. Section~\ref{sec:GATED_X--TFC_Formulation} develops \emph{Gated X--TFC} with three parts:  the forward solver,  the inverse parameter estimation, and  operator–conditioned meta learning. Section~\ref{sec:Results} reports results on the convection–diffusion benchmark with baseline comparisons followed by extension to twin boundary layer (multiple-subdomains) equation and a 2D Poisson equation with a sharp Gaussian source term. Section~\ref{sec:Conclusion} concludes the paper.

\section{Brief Recap of PINN, PIELM and X--TFC}
\label{sec:X--TFC_Review}
\paragraph{\textbf{PINNs}} In a typical Physics-Informed Neural Network (PINN) framework \citep{RAISSI2019686,JAGTAP2020113028,DWIVEDI2021299}, a deep neural network approximates the solution of PDEs. Randomly distributed collocation points within the computational domain, along with boundary points, serve as the training dataset. At these points, penalties are introduced for deviations from the governing PDE and boundary conditions (BCs). The mean squared error of these penalties, representing the residuals of the PDE and BCs, defines a physics-based loss function. This loss is minimized using gradient descent-based optimization algorithms and backpropagation. The core idea behind PINNs and their variants is to reformulate the task of solving PDEs as an optimization problem driven by a physics-informed loss function.

\paragraph{\textbf{PIELMs}}Physics-Informed Extreme Learning Machine (PIELM)\citep{DWIVEDI202096} differs from PINN in two key aspects: (A) network architecture and (B) cost function minimization. Unlike the deep multilayer architecture of PINNs, PIELM employs a single-hidden-layer design. A distinctive feature of PIELM is that its input layer weights are randomly initialized and remain fixed during training. With the first layer weights held constant, the PDE solution becomes linear with respect to the output layer weights, making the approximation process inherently linear. Instead of minimizing a physics-informed loss function through computationally intensive gradient descent and backpropagation techniques, PIELM formulates a system of linear residual equations, which can be efficiently solved using matrix inversion. 

\paragraph{\textbf{X--TFC}} The Extreme Theory of Functional Connections (X--TFC) \citep{SCHIASSI2021334} constructs an constrained trial function that satisfies boundary conditions exactly by design, eliminating boundary-penalty terms. Let $\mathcal{B}[u]=b$ denote the boundary operator and data. X-TFC derives a closed-form \emph{constrained expression}
\[
u(x) = g(x) + H(x)\bm{c},
\]
such that $\mathcal{B}[g]=b$ and $\mathcal{B}[H(\cdot)\bm{c}]=0$ for all coefficient vectors $\bm{c}$. Here, $g(x)$ enforces the constraints, while the columns of $H(x)$ span a basis of functions in the null space of $\mathcal{B}[\cdot]$. Choosing a linear dictionary for the free part (e.g., polynomials, Fourier/RBF features, or single-hidden-layer random features) makes the solution linear in $\bm{c}$. The PDE is then enforced only in the interior by minimizing a single residual least-squares objective
\[
\min_{\bm{c}} \sum_{j=1}^{N_{f}} \left\| \mathcal{L}\left[ g(x_{j}) + H(x_{j})\bm{c} \right] - f(x_{j}) \right\|^{2} + \eta \|\bm{c}\|^{2}.
\]
X--TFC thus (i) guarantees exact satisfaction of boundary conditions, (ii) avoids multi-objective tuning between PDE and BC losses, and (iii) trains quickly when $u$ is linear in $\bm{c}$.

\section{Gated X-TFC Formulation}
\label{sec:GATED_X--TFC_Formulation}
\subsection{Model problem}
We consider the one-dimensional steady convection--diffusion problem
\begin{equation}
	\label{eq:1d_ode}
	\mathcal{L}_{\nu}[u](x) := u'(x) - \nu u''(x) = 0,
	\quad x \in (0,1),
\end{equation}
subject to Dirichlet boundary conditions
\[
u(0) = B_L= 0, 
\qquad 
u(1) = B_R = 1,
\]
where $\nu > 0$ denotes the viscosity parameter. The closed-form solution to \eqref{eq:1d_ode} is
\begin{equation}
	\label{eq:1d_ode_sol}
	u(x;\nu) = \frac{e^{x/\nu} - 1}{e^{1/\nu} - 1}.
\end{equation}
When $\nu$ is small, i.e.\ $\nu \ll 1$, the solution develops a pronounced boundary layer close to the outflow boundary at $x=1$.

\subsection{Forward Problem Setup}
\begin{itemize}
	\item \textit{Trial Solution.} We use the standard X--TFC constrained expression: a global trial function that satisfies the Dirichlet boundary conditions exactly as follows:
	\[
	u(x;\bm c)=g(x)+\sum_{i=1}^{N_\star} c_i\,\psi_i(x),\qquad
	g(x)=(1-x)\,B_L+x\,B_R,\qquad \psi_i(0)=\psi_i(1)=0.
	\]
	We adopt Gaussian RBF features $\phi(z)=\exp(-z^2)$ with affine arguments
	$z_i(x)=m_i x+b_i$. The constrained basis functions are
	\[
	\psi_i(x)=\phi\!\big(z_i(x)\big)-(1-x)\,\phi\!\big(z_i(0)\big)-x\,\phi\!\big(z_i(1)\big),
	\]
	and the resulting convection-diffusion operators are given by
	\[
	\mathcal{L}_\nu[\psi_i](x)\;=\;\big(\phi(b_i)-\phi(m_i+b_i)\big)\;+\;m_i\,\phi'\!\big(z_i(x)\big)\;-\;\nu\,m_i^2\,\phi''\!\big(z_i(x)\big),
	\]
	and \(\mathcal{L}_\nu[g](x)=B_R-B_L\), a constant.
	\item \textit{RBF Centers Distribution.} Let \(x_s\in(0,1)\) split the domain into \([0,x_s)\) and \([x_s,1]\). We place \(N_c\) collocation points per block (\(N_f=2N_c\)) and \(N_\star\) RBF centers per block (\(N_s=2N_\star\)), uniformly within each block using endpoint grids:
	\[
	x_k^{\mathrm L}=x_L+\frac{k-1}{N_c}\,(x_s-x_L),\quad k=1,\dots,N_c,
	\qquad
	x_k^{\mathrm R}=x_s+\frac{k-1}{N_c}\,(x_R-x_s),\quad k=1,\dots,N_c,
	\]
	\[
	\alpha_j^{\mathrm L}=x_L+\frac{j-1}{N_\star}\,(x_s-x_L),\quad j=1,\dots,N_\star,
	\qquad
	\alpha_j^{\mathrm R}=x_s+\frac{j-1}{N_\star}\,(x_R-x_s),\quad j=1,\dots,N_\star.
	\]
	The affine parameters follow
	\[
	m_i=\frac{1}{\sqrt{2}\,\sigma_{x,i}},\qquad b_i=-m_i\,\alpha_i.
	\] 
	Thus the smaller block has smaller spacing and higher density. When $x_s=\tfrac12$ the layout is uniform on $[0,1]$; as $x_s\to 0$ or $x_s\to 1$ the design approaches a Shishkin-type mesh that clusters points near one boundary which is useful for resolving boundary layers.    
	\item \textit{RBF Widths Distribution.} The nominal left/right widths scale with the local center spacing:
	\[
	d_L=\frac{x_s}{N_\star},\quad d_R=\frac{1-x_s}{N_\star},\qquad
	\sigma_L=k\,d_L,\quad \sigma_R=k\,d_R,
	\]
	with a user-chosen multiplier $k>0$. To avoid a discontinuous jump at $x_s$, we blend these widths across the interface using a logistic gate. The transition scale $\varepsilon$ is chosen to be the maximum of a geometric scale, $\varepsilon_{\text{geo}} \propto \min(d_L, d_R)$, and a physical scale, $\varepsilon_{\text{phy}} \propto \nu$ (where $\nu$ is the diffusion / singular perturbation parameter, see \ref{app1} for details), ensuring the blend is informed by both the numerical discretization and the underlying physics of the problem. The blended width for each center $\alpha_i$ is then given by:
	\[
	s_i = \sigma\!\left( \frac{\alpha_i - x_s}{\varepsilon} \right) = \frac{1}{1 + \exp\left(-(\alpha_i - x_s)/\varepsilon\right)}, \qquad
	\sigma_{x,i} = (1 - s_i) \sigma_L + s_i \sigma_R.
	\]
	This approach ensures a smooth transition that adapts to both the mesh geometry and the physical characteristics of the boundary layer.
	\item \textit{Loss Function.} For the steady convection--diffusion operator \(\mathcal{L}_\nu[u]=u'-\nu u''\), the PDE residual of the X--TFC trial \(u(x;\boldsymbol{c})=g(x)+\sum_i c_i\psi_i(x)\) at collocation sites \(\{x_k\}_{k=1}^{N_f}\) is
	\[
	r_k(\boldsymbol{c};x_s,\varepsilon)\;=\;\mathcal{L}_\nu[u](x_k)
	=\underbrace{\mathcal{L}_\nu[g](x_k)}_{f_k}\;+\;\sum_{i=1}^{N_s} c_i\,\underbrace{\mathcal{L}_\nu[\psi_i](x_k)}_{R_{k i}},
	\quad k=1,\dots,N_f.
	\]
	We minimize a weighted least-squares residual with a small Tikhonov term:
	\[
	\mathcal{J}(\boldsymbol{c};x_s,\varepsilon)
	=\sum_{k=1}^{N_f} w_k\,r_k(\boldsymbol{c};x_s,\varepsilon)^2\;+\;\lambda\|\boldsymbol{c}\|_2^2
	=\big\|R(x_s,\varepsilon)\,\boldsymbol{c}+ \boldsymbol{f}\big\|_{W}^2+\lambda\|\boldsymbol{c}\|_2^2,
	\]
	where \(W=\mathrm{diag}(w_1,\dots,w_{N_f})=(1/N_f)\mathbf{1}\), \(\lambda\ge0\), \(\boldsymbol{f}=[f_k]\), and \(R=[R_{k i}]\). The dependence on \(x_s\) and \(\varepsilon\) enters only through the centers $\alpha_{i}$ and widths $\sigma_{x,i}$ defined by the soft split and logistic gating defined earlier.
	\item \textit{Optimization.} 
	\begin{enumerate}
		\item \textit{Inner Optimization.}     For fixed \((x_s,\varepsilon)\), the coefficient vector is obtained by a linear least-squares solve
		\[
		\boldsymbol{c}^\star(x_s,\varepsilon)\;=\;\arg\min_{\boldsymbol{c}}\;\mathcal{J}(\boldsymbol{c};x_s,\varepsilon)
		\quad\Longleftrightarrow\quad
		\big(R^\top W R+\lambda I\big)\,\boldsymbol{c}^\star=-R^\top W\,\boldsymbol{f},
		\]
		solved stably via Cholesky or QR.
		\item \textit{Outer Optimization.} The hyperparameters are then selected by a simple outer search:
		\[
		(x_s^\star,\varepsilon^\star)
		=\arg\min_{x_s\in(0,1),\,\varepsilon>0}\;\mathcal{E}\big(u(\,\cdot\,;\boldsymbol{c}^\star(x_s,\varepsilon)\big),
		\]
		where the outer objective \(\mathcal{E}(\cdot)\) is residual energy on an independent validation grid. In all cases, no interface loss terms are introduced; continuity across the split is maintained by the global expansion and the smooth width gating. Please note that this forward solve does \emph{not} use Bayesian optimization; outer hyperparameters are selected with deterministic ridge-based nested \texttt{fminbnd} \footnote{https://www.mathworks.com/help/matlab/ref/fminbnd.html} on a validation grid.
	\end{enumerate}
	
\end{itemize}
The split \(x_s\) sets \emph{where} resolution is concentrated, while the transition scale \(\varepsilon\) controls \emph{how} quickly kernel widths vary across that location; both are important for resolving thin layers without ill conditioning. Although we describe one split and one transition scale for clarity, extending to multiple interior splits is straightforward by composing the same partition-and-gate construction.

\subsection{Inverse Problem Setup}
We consider the same 1D convection--diffusion model (Eq.~\ref{eq:1d_ode}) but treat the diffusion parameter \(\nu>0\) as \emph{unknown}, to be inferred from noisy pointwise observations
\[
y_n \;=\; u(x_n;\nu) \;+\; \varepsilon_n,\qquad \varepsilon_n \sim \mathcal{N}(0,\sigma^2),\quad n=1,\dots,N_{\text{data}}.
\]
For controlled experiments we synthesize data from the analytic solution and add Gaussian noise with known standard deviation \(\sigma\). To better resolve boundary layers when \(\nu\ll 1\), both the data sites \(\{x_n\}\) and the PDE collocation points are right–clustered toward \(x=1\).

\begin{itemize}
	\item \textit{Trial solution, RBF Centers, and Widths.} We reuse the forward X--TFC constrained expression,
	\[
	u(x;\boldsymbol{c}) \;=\; g(x)\;+\;\sum_{i=1}^{N_s} c_i\,\psi_i(x), 
	\qquad g(x)=(1-x)B_L + x B_R,
	\]
	with Gaussian features \(\phi(z)=e^{-z^2}\) evaluated at affine arguments \(z_i(x)=m_i x + b_i\). The constrained basis functions \(\psi_i\) vanish at the boundaries and are built from \(\phi\) as in the forward section; centers \(\{\alpha_i\}_{i=1}^{N_s}\) and widths \(\{\sigma_{x,i}\}_{i=1}^{N_s}\) are induced by a soft split at \(x_s\) with a logistic gate of transition scale \(\varepsilon\). 
	
	\item \textit{Whitened Linear Model (Data + Physics).}
	Let \(H_{\text{data}}\in\mathbb{R}^{N_{\text{data}}\times N_s}\) collect the rows \([H_{\text{data}}]_{n i}=\psi_i(x_n)\) for \(n=1,\dots,N_{\text{data}}\), and define the data right–hand side after removing the constrained part,
	\[
	\boldsymbol{y}_{\text{data}} \;=\; \bigl[y_n - g(x_n)\bigr]_{n=1}^{N_{\text{data}}}.
	\]
	Let \(R\in\mathbb{R}^{N_f\times N_s}\) collect strong–form residual rows \([R]_{k i}=\mathcal{L}_\nu[\psi_i](x_k)\) at PDE collocation sites \(\{x_k\}_{k=1}^{N_f}\), and set \(\boldsymbol{f}=\bigl[\mathcal{L}_\nu[g](x_k)\bigr]_{k=1}^{N_f}=(B_R-B_L)\,\mathbf{1}\). With data precision \(\beta_{\text{data}}=1/\sigma^2\) and PDE precision \(\beta_{\text{pde}}>0\) (both known constants), we \emph{whiten} and stack the two blocks:
	\[
	\Phi\,\boldsymbol{c}\;\approx\;\boldsymbol{y},\qquad
	\Phi=\begin{bmatrix}
		\sqrt{\beta_{\text{data}}}\,H_{\text{data}}\\[3pt]
		\sqrt{\beta_{\text{pde}}}\,R
	\end{bmatrix}\in\mathbb{R}^{N\times N_s},\qquad
	\boldsymbol{y}=\begin{bmatrix}
		\sqrt{\beta_{\text{data}}}\,\boldsymbol{y}_{\text{data}}\\[3pt]
		-\sqrt{\beta_{\text{pde}}}\,\boldsymbol{f}
	\end{bmatrix}\in\mathbb{R}^{N},
	\]
	where \(N=N_{\text{data}}+N_f\). Note that \(R\) (hence \(\Phi\)) depends on \((\nu,x_s,\varepsilon)\) through \(\mathcal{L}_\nu\) and through the gated centers/widths. This yields a single linear system coupling data and physics, with \emph{no} boundary or interface penalties.
	
	\item \textit{Optimization.}
	\begin{enumerate}
		\item \textit{Inner Optimization.} We place a zero–mean isotropic Gaussian prior on the coefficients,
		\[
		\boldsymbol{c}\sim\mathcal{N}\!\bigl(\mathbf{0},\,\eta^{-1}\mathbf{I}\bigr),\qquad \eta>0,
		\]
		where \(\eta\) is the prior precision. For a given \((\nu,x_s,\varepsilon)\), the posterior over \(\boldsymbol{c}\) is 
		\[
		A(\eta,\nu,x_s,\varepsilon))\;=\;\eta\mathbf{I}+\Phi^\top\Phi,\qquad
		\boldsymbol{m}_N(\eta,\nu,x_s,\varepsilon))\;=\;A^{-1}\Phi^\top\boldsymbol{y}.
		\]
		The marginal log-likelihood (evidence) (Chapter 3 of PMRL book \cite{bishop2006pattern}) is
		\[
		\log p(\boldsymbol{y}\mid \eta,\nu,x_s,\varepsilon)
		=\frac{M}{2}\log\eta
		-\frac{1}{2}\bigl(\|\boldsymbol{y}-\Phi\boldsymbol{m}_N\|_2^2+\eta\|\boldsymbol{m}_N\|_2^2\bigr)
		-\frac{1}{2}\log|A(\eta)|-\frac{N}{2}\log(2\pi),
		\]
		with $M=N_s$ the number of basis functions and $N$ the number of stacked rows. For any candidate \((\nu,x_s,\varepsilon)\), we optimize the coefficient prior precision $\eta$ by a damped fixed-point update initialized at a user-specified $\eta_0=10^{-7}$ and constrained to $[\eta_{\min},\eta_{\max}]=[10^{-12},10^{-2}]$:
		\[
		\gamma=\sum_{i=1}^{M}\frac{\kappa_i^2}{\kappa_i^2+\eta},\qquad
		\eta\leftarrow \frac{\gamma}{\|m_N\|_2^2}.
		\]
		where $\{\kappa_i^2\}$ are the eigenvalues of $\Phi^\top\Phi$.
		
		\item \textit{Outer Optimization.} We then maximize the evidence over the hyperparameters,
		\[
		(\nu^\star,x_s^\star,\varepsilon^\star)
		\;=\;\arg\max_{\nu>0,\;x_s\in(0,1),\;\varepsilon>0}\;\;\max_{\eta>0}\;\log p(\boldsymbol{y}\mid \eta^\star,\nu,x_s,\varepsilon),
		\]
		using derivative–free Bayesian optimization over \((\nu,x_s,\varepsilon)\). Each BO evaluation rebuilds \(\Phi\), re–optimizes \(\eta\), and returns the \emph{negative} log–evidence as a scalar objective for minimization.
	\end{enumerate}
	\item \textit{Posterior prediction.} With \((\nu^\star,x_s^\star,\varepsilon^\star,\eta^\star)\) fixed, the posterior mean and (model) predictive variance at any \(x\in[0,1]\) are
	\[
	u_{\text{mean}}(x)\;=\;g(x)+\boldsymbol{h}(x)^\top\boldsymbol{m}_N,\qquad
	\sigma_{\text{model}}^2(x)\;=\;\boldsymbol{h}(x)^\top A(\eta^\star)^{-1}\boldsymbol{h}(x),
	\]
	where \(\boldsymbol{h}(x)=\bigl[\psi_1(x),\dots,\psi_{N_s}(x)\bigr]^\top\). These yield calibrated uncertainty bands for the reconstructed field.
	
\end{itemize}

In summary, we select \(\eta\) by evidence and maximize the evidence over \((\nu, x_s, \varepsilon)\) using Bayesian optimization; each evaluation rebuilds \(\Phi\) and re-optimizes \(\eta\). As in the forward case, no interface penalties are used; the soft split governs where (via \(x_s\)) and how (via \(\varepsilon\)) the kernel resolution adapts.
\subsection{Operator–Conditioned Meta–Learning}
\label{sec:meta}
Given a differential operator parameter (denoted here as $\nu$), our objective is to \emph{predict} effective gating hyperparameters for the Gated X--TFC method—specifically, the split location $x_s$ and the transition scale $\varepsilon_{\text{scale}}$. This prediction aims to initialize the subsequent optimization near a promising region, enabling faster convergence and reducing the number of function evaluations in both forward and inverse problems. The procedure consists of five main steps:

\begin{enumerate}
	\item \textit{Data generation.} We begin by sampling the diffusion parameter across a desired range. For each sampled parameter, the forward Gated X--TFC solver is executed to obtain the optimal gate parameters (split location and transition scale) that minimize the physics residual the most effectively. These pairs of operator parameters and their corresponding optimal gates create a compact supervised dataset
	\[
	\mathcal{D} = \big\{\,(\nu_j, \; x_{s,j}^*, \; \varepsilon_j^*)\,\big\}_{j=1}^{N_{\rm train}},
	\]
	which is then employed to learn a predictive mapping from the operator parameter to the recommended gate configurations.
	
	\item \textit{Target Variable Transformation.} To ensure stable regression, we apply monotonic transformations to map constrained parameters onto an $\mathcal{O}(1)$ scale. Specifically, we use $\log_{10}\nu$ for the diffusion parameter, a logit function for the split (scaled between [0.8,0.999]), and $\log_{10}\varepsilon_{\text{scale}}$ for the gate scale. These transformations simplify the learning process while guaranteeing that inverted outputs remain within valid physical ranges.
	
	\item \textit{Heteroskedastic Regression.} We observe that the relationship between the operator and the optimal gates is not uniformly precise; variability in the optimal settings is greater for smaller diffusion values, where solutions develop sharper layers. To account for this uneven noise (heteroskedasticity), we use a regularized regression model that incorporates data-dependent weights. These weights are informed by the local spread of the training residuals, allowing the model to produce a stable global fit that accurately captures smooth trends while also quantifying regions of higher uncertainty.
	
	\item \textit{Prediction and Uncertainty Quantification.} For a new, unseen value of the diffusion parameter, the trained meta-model provides a prediction in the transformed space. This prediction includes both an expected value and an estimate of uncertainty. We then apply the inverse transformations to convert these outputs back into physically meaningful gate settings and their associated confidence intervals. These intervals provide a crucial measure of the model's confidence in its own recommendations.
	
	\item \textit{Informed Optimization at Test Time.} During deployment, the model's predictions are used to highly accelerate the outer optimizer. The predicted mean values serve as intelligent warm-start initializations. The confidence bands define narrow, instance-specific search boxes around these starts, drastically reducing the domain the optimizer must explore. If the model indicates high uncertainty for a particular input, these search windows can be automatically expanded toward the original global bounds. This strategy of conditioning the optimizer on the operator value reduces the number of costly evaluations required, serving as an efficient alternative to building a full surrogate model for the entire operator. It integrates directly and seamlessly with the existing Gated X–TFC framework.    
\end{enumerate}
This procedure amounts to learning a conditioning policy that maps the operator parameter to solver hyperparameters. It is not a surrogate of the PDE solution itself; instead, it meta–learns the gate configuration that makes the downstream physics solve cheap and stable. In this sense it provides a light–weight, data–efficient alternative to full operator surrogates, and integrates seamlessly with the Gated X--TFC training loop.
\section{Results and Discussion}
\label{sec:Results}
This section evaluates Gated X–TFC on the 1D convection–diffusion benchmark (Eq.~\ref{eq:1d_ode}) across regimes from smooth solutions to sharp boundary layers. We assess interpretability (learned gates and width profiles), forward accuracy, inverse parameter recovery, and operator-conditioned meta-learning. We also compare our method against state-of-the-art ungated X–TFC, PIELM, Deep-TFC and a baseline PINN. Finally, we demonstrate our method's applicability in twin boundary layer (multiple-subdomains) equation  and 2D Poisson equation with a sharp source term.  

All experiments were run in \textsc{Matlab} R2022b on a 12th-Gen Intel\textsuperscript{\textregistered} Core\texttrademark{} i7-12700H (2.30 GHz) with 16 GB RAM on an ASUS laptop.

\subsection{Forward Solution}
\label{sec:results-forward}
\begin{figure}[htbp]
	\centering
	\includegraphics[width=0.8\linewidth]{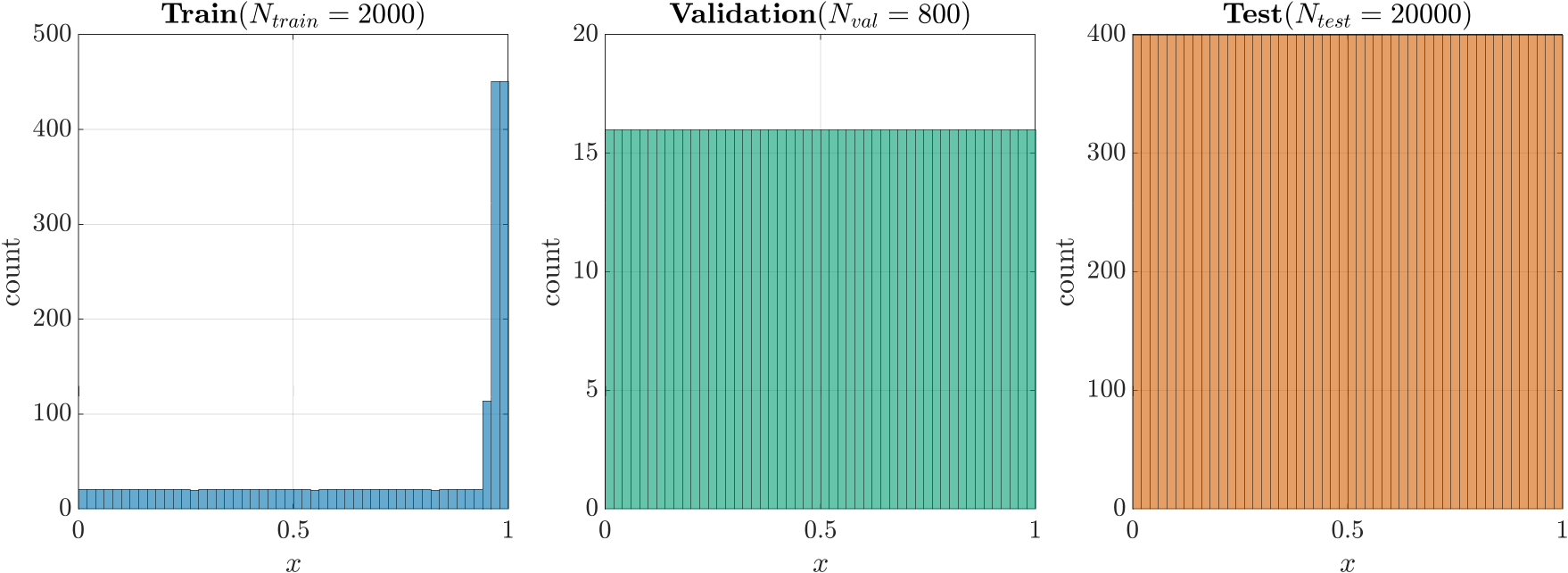} 
	\caption{Distribution of training, validation and test collocation points.}
	\label{fig:train-val-test}
\end{figure}

\begin{figure}[htbp]
	\centering
	\begin{subfigure}[t]{0.38\textwidth}
		\centering
		\includegraphics[width=\linewidth]{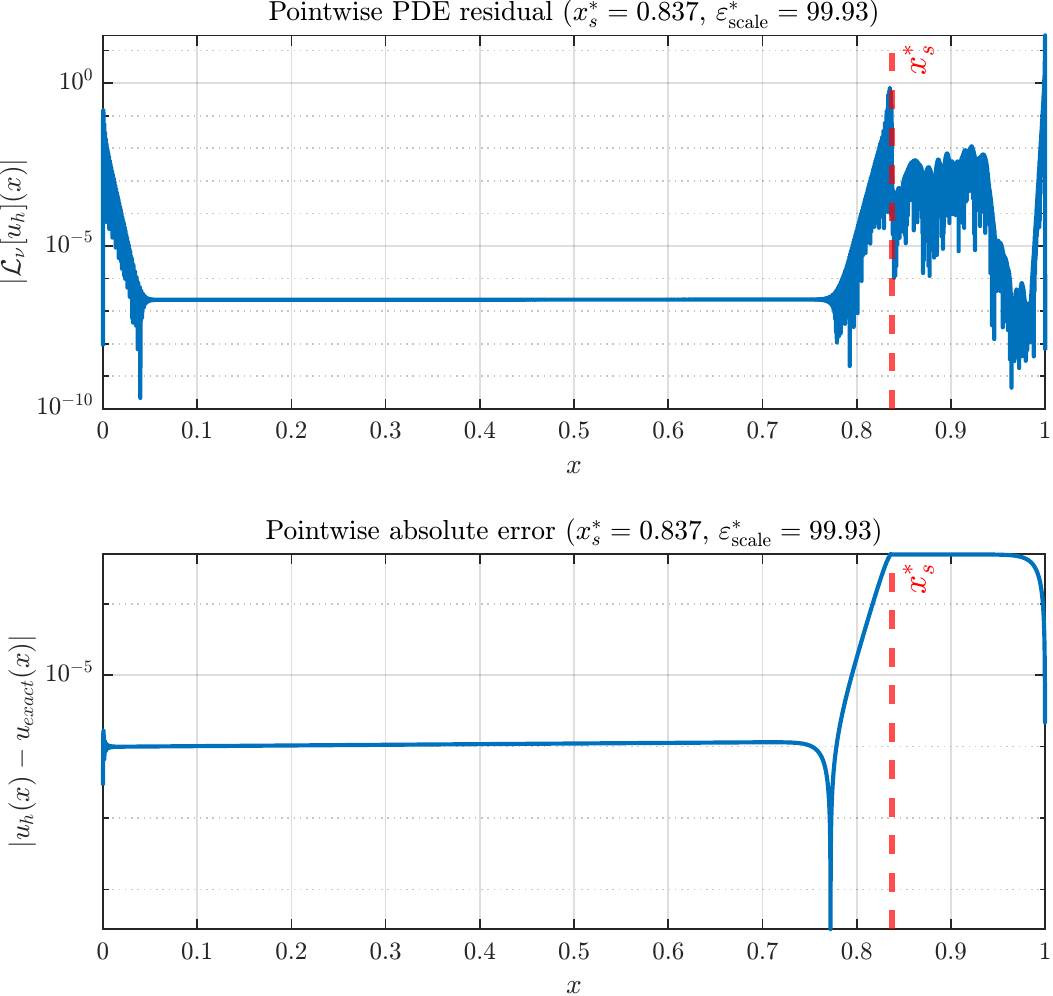}
		\caption{Residual \(\mathcal{L}_\nu[u_h]\) and absolute error, \(\nu=10^{-2}\). Dashed line: \(x_s^\ast\).}
		\label{fig:fwd_1e-2_ptwise}
	\end{subfigure}\hfill
	\begin{subfigure}[t]{0.38\textwidth}
		\centering
		\includegraphics[width=\linewidth]{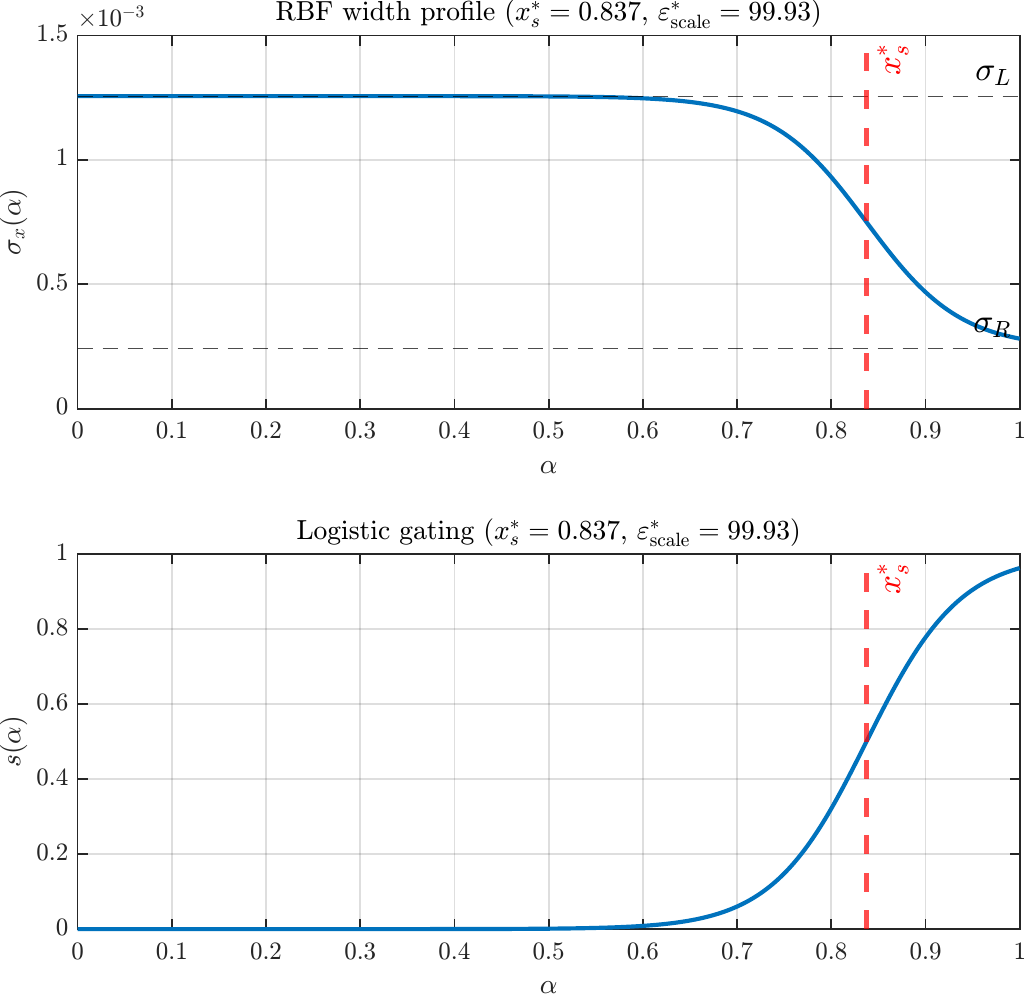}
		\caption{Width \(\sigma_x(\alpha)\) and gate \(s(\alpha)\), \(\nu=10^{-2}\). Transition scale \(\varepsilon_{\mathrm{scale}}^\ast\).}
		\label{fig:fwd_1e-2_gate}
	\end{subfigure}
	
	\medskip
	
	\begin{subfigure}[t]{0.38\textwidth}
		\centering
		\includegraphics[width=\linewidth]{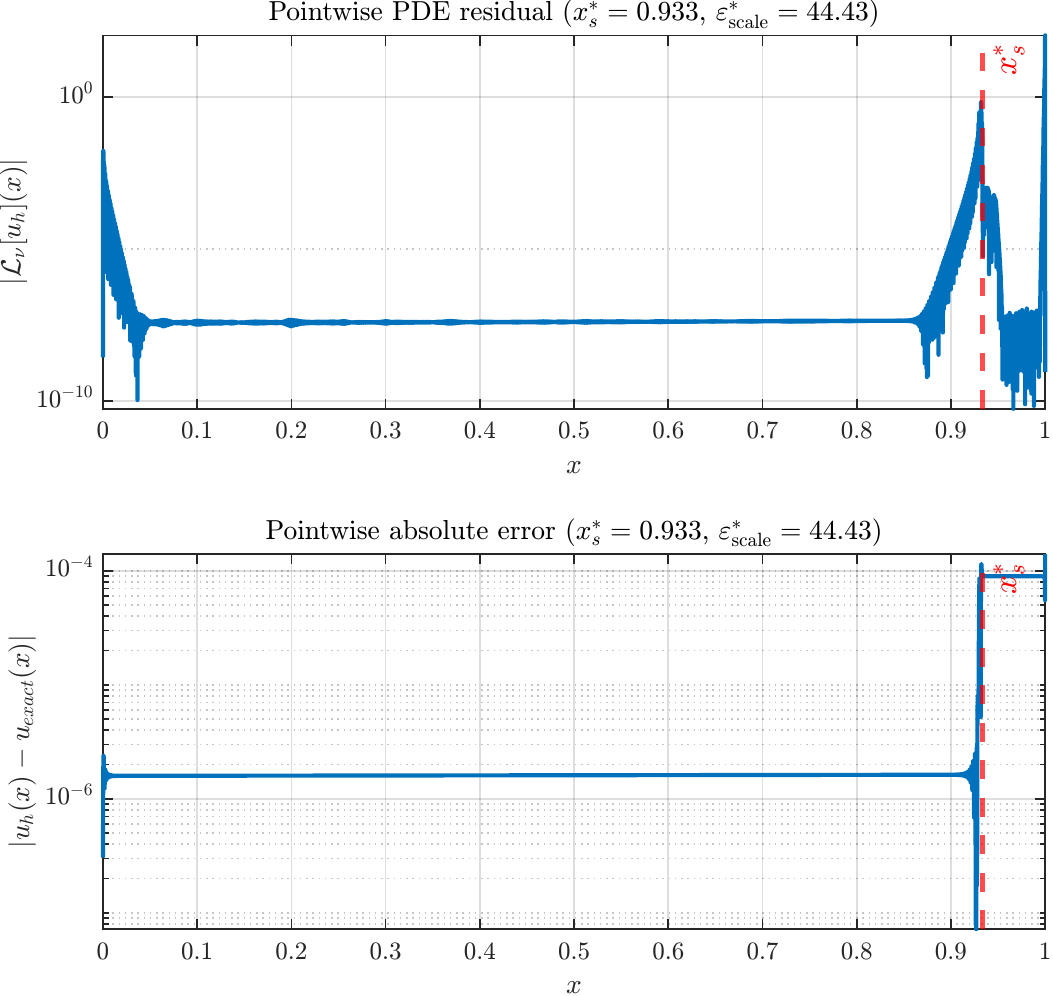}
		\caption{Residual and absolute error, \(\nu=10^{-3}\). Dashed line: \(x_s^\ast\).}
		\label{fig:fwd_1e-3_ptwise}
	\end{subfigure}\hfill
	\begin{subfigure}[t]{0.38\textwidth}
		\centering
		\includegraphics[width=\linewidth]{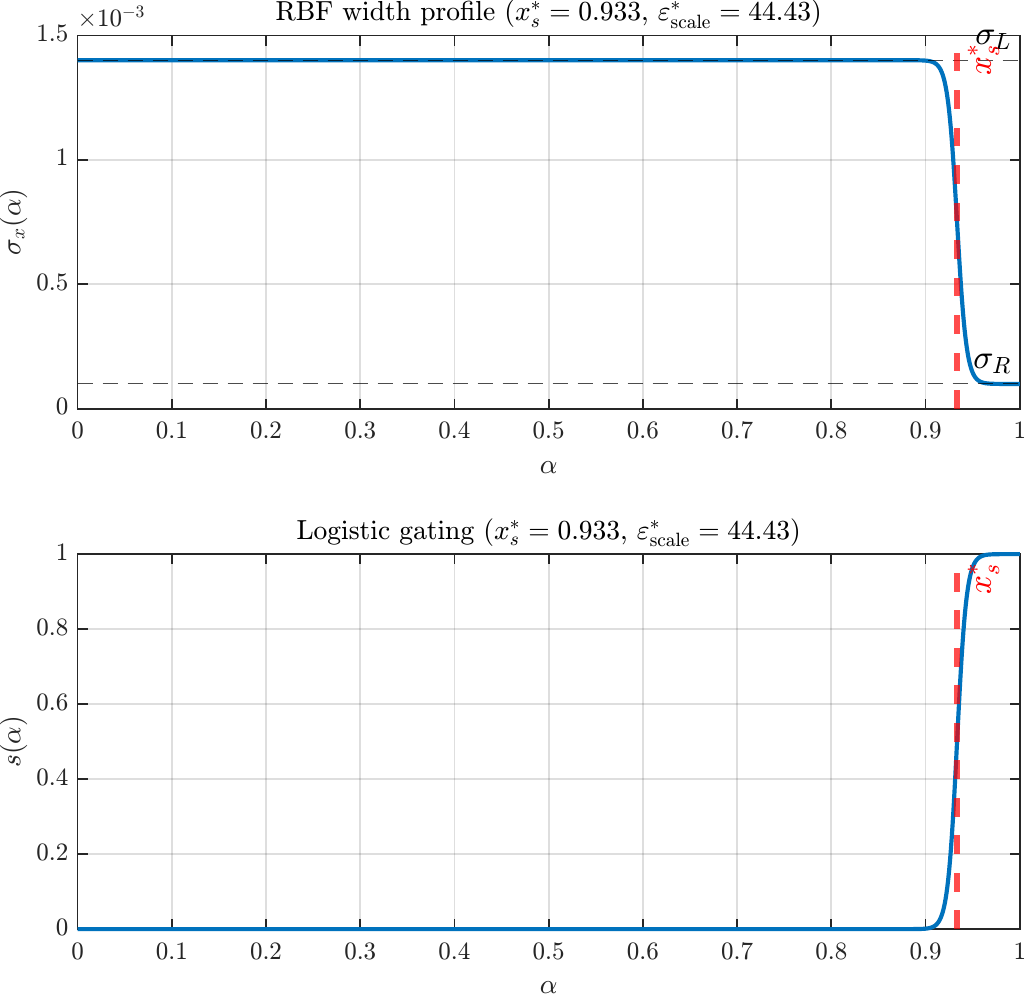}
		\caption{Width and gate profiles, \(\nu=10^{-3}\). Transition scale \(\varepsilon_{\mathrm{scale}}^\ast\).}
		\label{fig:fwd_1e-3_gate}
	\end{subfigure}
	
	\medskip
	
	\begin{subfigure}[t]{0.38\textwidth}
		\centering
		\includegraphics[width=\linewidth]{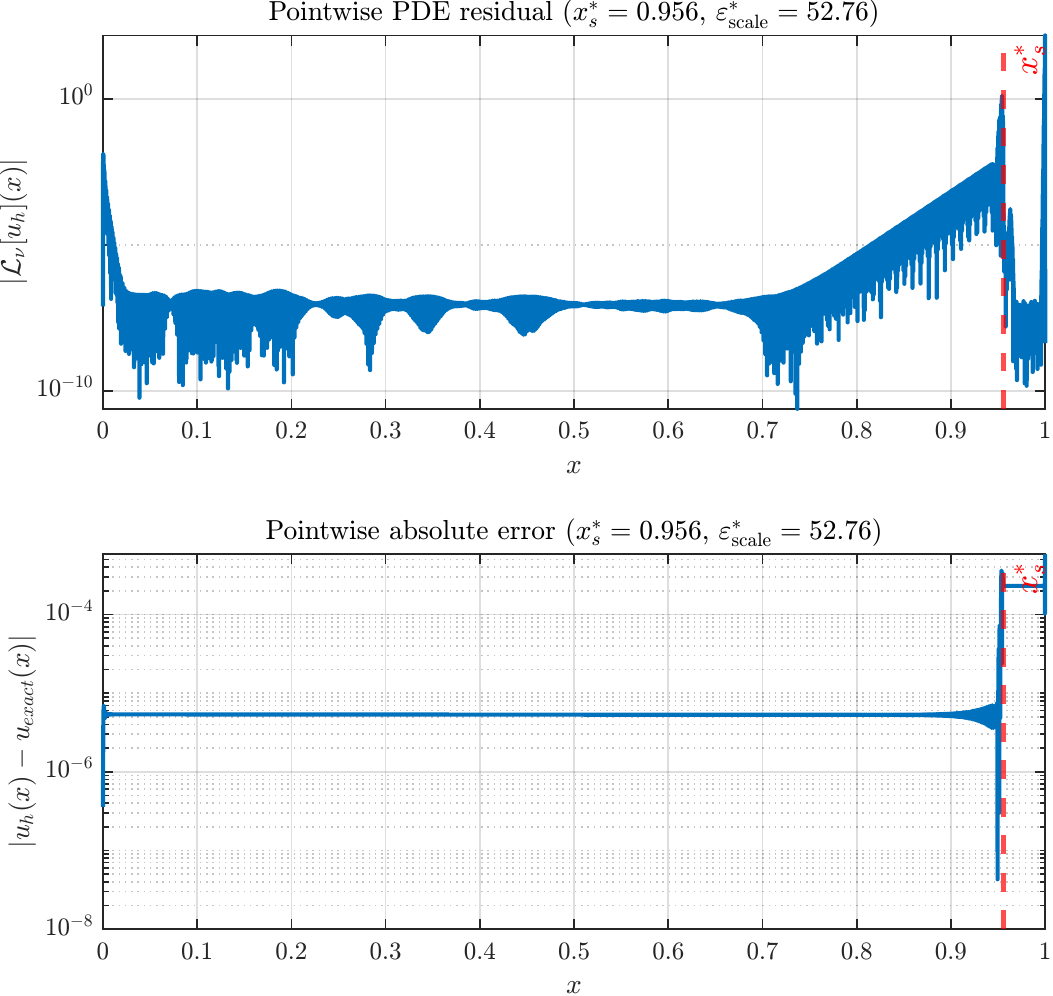}
		\caption{Residual and absolute error, \(\nu=10^{-4}\). Dashed line: \(x_s^\ast\).}
		\label{fig:fwd_1e-4_ptwise}
	\end{subfigure}\hfill
	\begin{subfigure}[t]{0.38\textwidth}
		\centering
		\includegraphics[width=\linewidth]{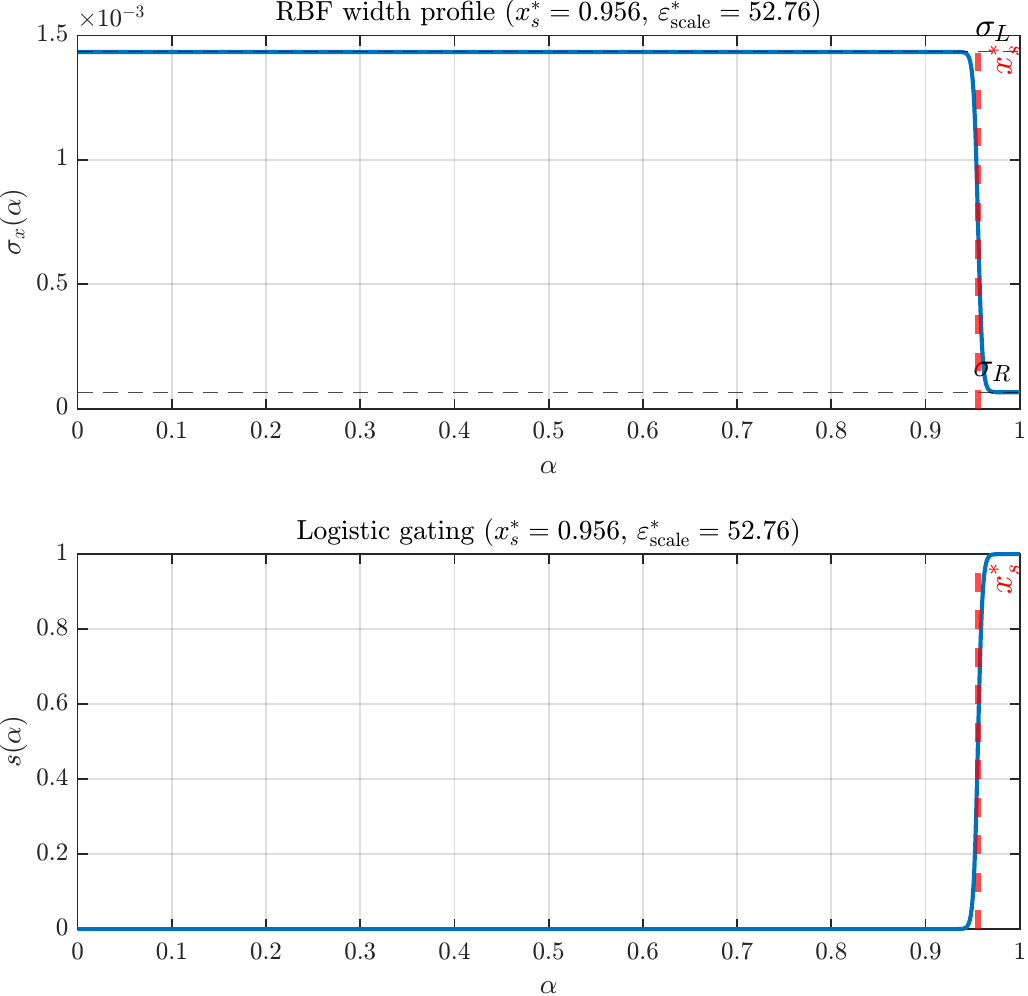}
		\caption{Width and gate profiles, \(\nu=10^{-4}\). Transition scale \(\varepsilon_{\mathrm{scale}}^\ast\).}
		\label{fig:fwd_1e-4_gate}
	\end{subfigure}
	
	\caption{Gated X--TFC across decreasing diffusion. Left column: optimization target (pointwise PDE residual) and absolute error shown only for assessment; no interface penalties are used. Right column: learned RBF width field and logistic gate; the dashed vertical line marks the learned split \(x_s^\ast\).}
	\label{fig:gated_xtfc_grid}
\end{figure}
\begin{figure}[htbp]
	\centering
	\begin{subfigure}{0.7\textwidth}
		\centering
		\includegraphics[width=\linewidth]{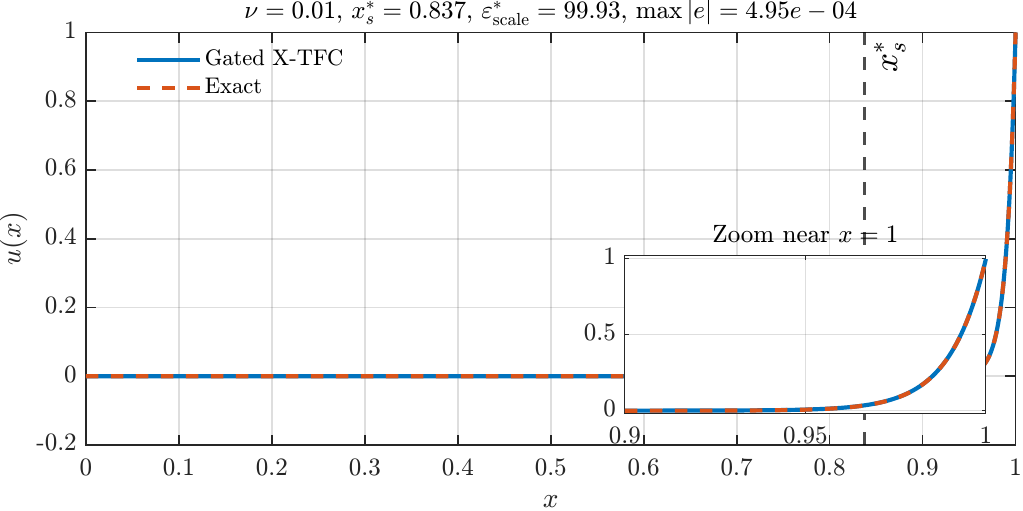}
		\caption{Gated X--TFC vs exact at $\nu=10^{-2}$}
	\end{subfigure}
	
	\begin{subfigure}{0.7\textwidth}
		\centering
		\includegraphics[width=\linewidth]{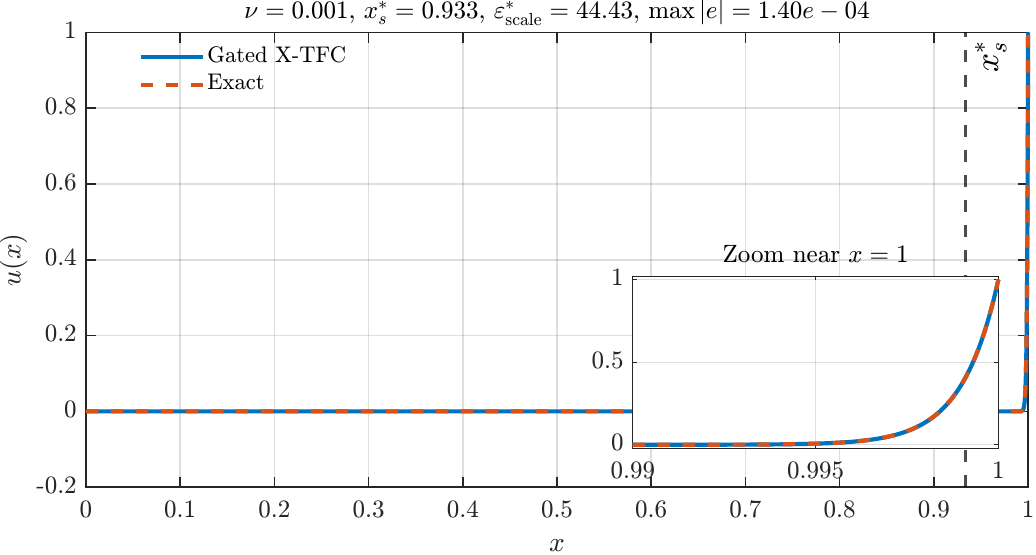}
		\caption{Gated X--TFC vs exact at $\nu=10^{-3}$}
	\end{subfigure}
	
	\begin{subfigure}{0.7\textwidth}
		\centering
		\includegraphics[width=\linewidth]{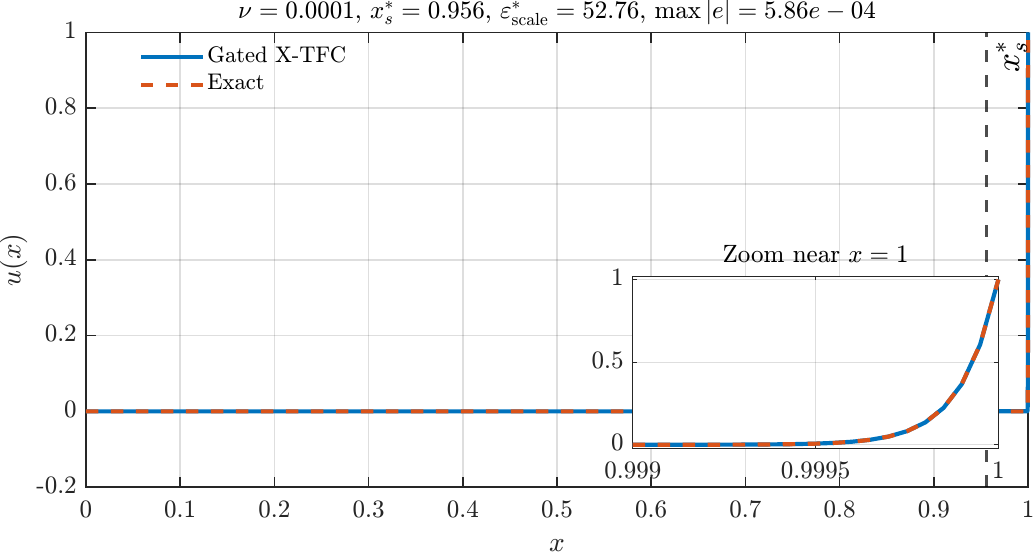}
		\caption{Gated X--TFC vs exact at $\nu=10^{-4}$}
	\end{subfigure}
	\caption{Gated X--TFC solution (solid) vs.\ exact (dashed) for $\nu=[10^{-2},10^{-3},10^{-4}]$. The dashed vertical line marks the learned split $x_s^\ast$; the inset zooms the boundary layer near $x=1$. The exact solution is used only for assessment—training minimizes the strong-form residual with no interface penalties.}
	\label{fig:fwd_comparisons}
\end{figure}
We evaluate the Gated X-TFC framework on the steady 1D convection--diffusion problem~\eqref{eq:1d_ode} across three decreasing diffusion regimes: $\nu \in \{10^{-2}, 10^{-3}, 10^{-4}\}$, maintaining a fixed numerical setup throughout. For each value of $\nu$, the split location $x_s^*$ and transition scale $\varepsilon$ are optimized by minimizing the strong-form residual, without employing any boundary or interface penalties. 

\paragraph{Experimental Configuration}
The following setup remains consistent across all values of $\nu$:
\begin{itemize}
	\item \textit{Collocation and Basis:} Two softly decomposed subdomains share a single global trial function. Each subdomain contains $N_c = 1000$ collocation points and $N_\star = 1000$ RBF centers, resulting in totals of $2N_c = 2000$ points and $2N_\star = 2000$ centers. Gaussian RBF widths are defined per subdomain as $\sigma_L = k \cdot d_L$ and $\sigma_R = k \cdot d_R$, with $k = 1.5$, and local grid spacings $d_L = x_s / N_\star$, $d_R = (1 - x_s) / N_\star$.
	
	\item \textit{Adaptive Gating Mechanism:} A logistic gate $s(\alpha) = \left(1 + \exp(-t)\right)^{-1}$, where $t = (\alpha - x_s)/\varepsilon_b$, blends the RBF widths smoothly:
	\begin{equation*}
		\sigma_x(\alpha) = (1 - s(\alpha)) \sigma_L + s(\alpha) \sigma_R
	\end{equation*}
	The transition width $\varepsilon_b$ is determined as $\varepsilon_b = \max(\varepsilon_{\text{geo}}, \varepsilon_{\text{phy}})$, where $\varepsilon_{\text{geo}} = \varepsilon_{\text{scale}} \cdot \min(d_L, d_R)$ is a geometric scale and $\varepsilon_{\text{phy}} = 5\nu$ provides a physics-informed lower bound.
	
	\item \textit{Bilevel Optimization:} The split location $x_s \in [0.80, 0.999]$ (outer variable) and the geometric scale $\varepsilon_{\text{scale}} \in [10, 100]$ (inner variable) are optimized using MATLAB's \texttt{fminbnd} with tolerances $\mathrm{TolX}{x_s} = 10^{-4}$ and $\mathrm{TolX}{\varepsilon} = 10^{-1}$, respectively.
	
	\item \textit{Objective and Evaluation:} Optimization minimizes the PDE residual on an independent validation set of $N_v = 800$ points to prevent overfitting. Performance is evaluated on a dense test grid of $N_{\text{test}} = 2 \times 10^4$ points, with the exact solution used only for visualization and error analysis (see Fig. \ref{fig:train-val-test}).
\end{itemize}
\paragraph{Results and Analysis}
Figure~\ref{fig:gated_xtfc_grid} presents pointwise residuals and absolute errors (left column), along with the adapted width and gate profiles (right column), for each value of $\nu$. As $\nu$ decreases, the optimized split location $x_s^*$ migrates toward the outflow boundary ($x = 1$), and the gate sharpens to concentrate numerical resolution within the narrowing boundary layer. The residual remains small throughout the domain, exhibiting localized peaks only near the interface. Absolute errors show similar localization.

Across all three $\nu$, the predicted solutions match the exact profiles accurately (Fig.~\ref{fig:fwd_comparisons}); inset zooms show no spurious oscillations near the boundary layer. Notably, this accuracy is achieved without any continuity or penalty terms at the interior split---the global X-TFC trial remains continuous and satisfies boundary conditions exactly by construction.

\paragraph{Comparison with State-of-the-Art}
\citet{DEFLORIO2024115396} conducted a comprehensive benchmark study evaluating four surrogate modeling approaches—classic Physics-Informed Neural Networks (PINNs), Physics-Informed Extreme Learning Machines (PIELM), Deep Theory of Functional Connections (Deep-TFC), and Extreme Theory of Functional Connections (X-TFC)—on the same 1D steady advection--diffusion problem with $\nu\in\{10^{-1},10^{-2},10^{-3},10^{-4}\}$. Their results demonstrate that ELM-based methods (PIELM and X-TFC) consistently achieve the lowest errors and training losses, with X-TFC performing best due to its exact enforcement of boundary conditions.

For the mild case $\nu=1$, all methods perform satisfactorily, though with varying precision: mean absolute errors are $\sim$$10^{-5}$ for PINNs, $\sim$$10^{-8}$ for Deep-TFC, $\sim$$10^{-15}$ for PIELM, and $\sim$$10^{-16}$ for X-TFC. As the diffusion coefficient decreases, performance diverges significantly. PINNs fail at $\nu=10^{-2}$, while Deep-TFC begins to show bias at this regime. At $\nu=10^{-3}$, Deep-TFC fails completely while PIELM and X-TFC maintain accuracies of $\sim$$10^{-8}$ and $\sim$$10^{-14}$, respectively. In the most challenging case ($\nu=10^{-4}$), only X-TFC remains viable, achieving a mean absolute error of $\sim$$10^{-3}$. The authors attribute the difficulty in this regime to poorly resolved derivatives as the solution approaches discontinuity, suggesting domain decomposition as a potential mitigation strategy—precisely the limitation our gated formulation addresses.

For the most challenging case $\nu=10^{-4}$, our Gated X-TFC method demonstrates decisive improvements over the best ungated X-TFC baseline in both accuracy and computational efficiency. Table~\ref{tab:gated_vs_xtfc} provides a detailed comparison under identical hardware and software configurations.

\begin{table}[htbp]
	\centering
	\small
	\caption{Performance comparison between X-TFC and Gated X-TFC at $\nu=10^{-4}$. The gated variant achieves an order-of-magnitude error reduction while using 80\% fewer points/neurons and approximately 66\% less computational time.}
	\label{tab:gated_vs_xtfc}
	\begin{tabular}{lcccc}
		\toprule
		\textbf{Method} & \textbf{Training Points} & \textbf{Neurons} & \textbf{Maximum Error} & \textbf{Training Time (s)} \\
		\midrule
		X-TFC (ungated) & $10{,}000$ & $10{,}000$ & $\mathcal{O}(10^{-3})$ & $153$ \\
		Gated X-TFC (ours) & $2{,}000$ & $2{,}000$ & $\mathcal{O}(10^{-4})$ & $52$ \\
		\midrule
		Improvement & $5\times$ fewer & $5\times$ fewer & $10\times$ lower & $\approx 2.9\times$ faster \\
		& ($-80\%$) & ($-80\%$) & & ($-66\%$) \\
		\bottomrule
	\end{tabular}
\end{table}
\subsection{Inverse Inference}
\label{sec:results-inverse}

\begin{figure}[htbp]
	\centering
	\begin{subfigure}{0.48\textwidth}
		\includegraphics[width=\linewidth]{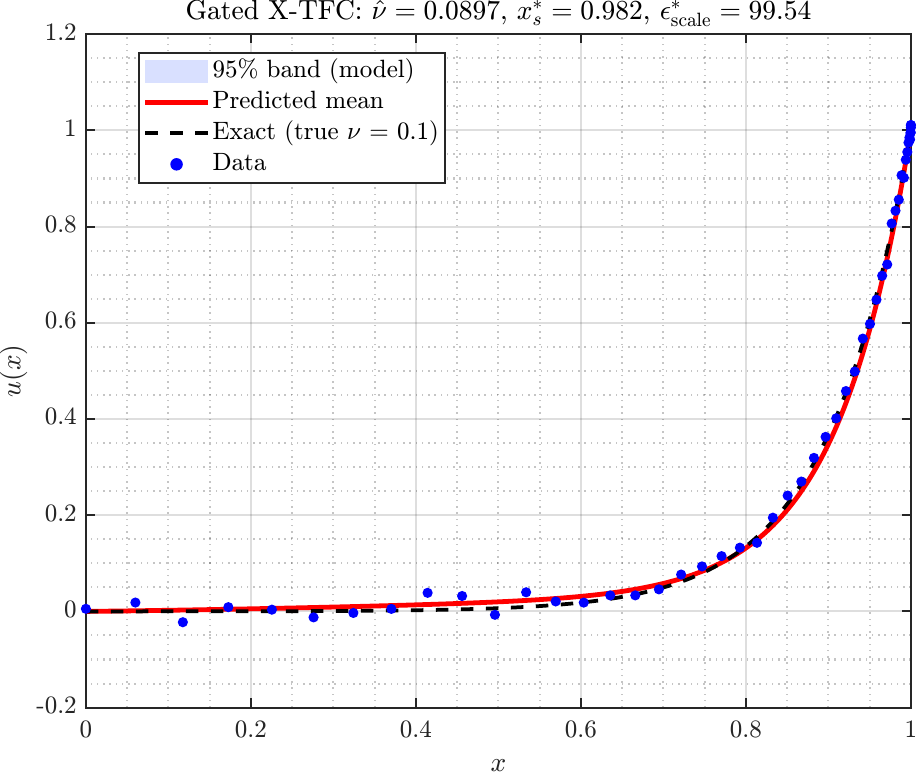}
		\caption{Inverse estimation at $\nu_{\text{true}}=10^{-1}$: mild layer; gate close to uniform, accurate $\hat{\nu}$ with a tight $95\%$ band.}
	\end{subfigure}\hfill
	\begin{subfigure}{0.48\textwidth}
		\includegraphics[width=\linewidth]{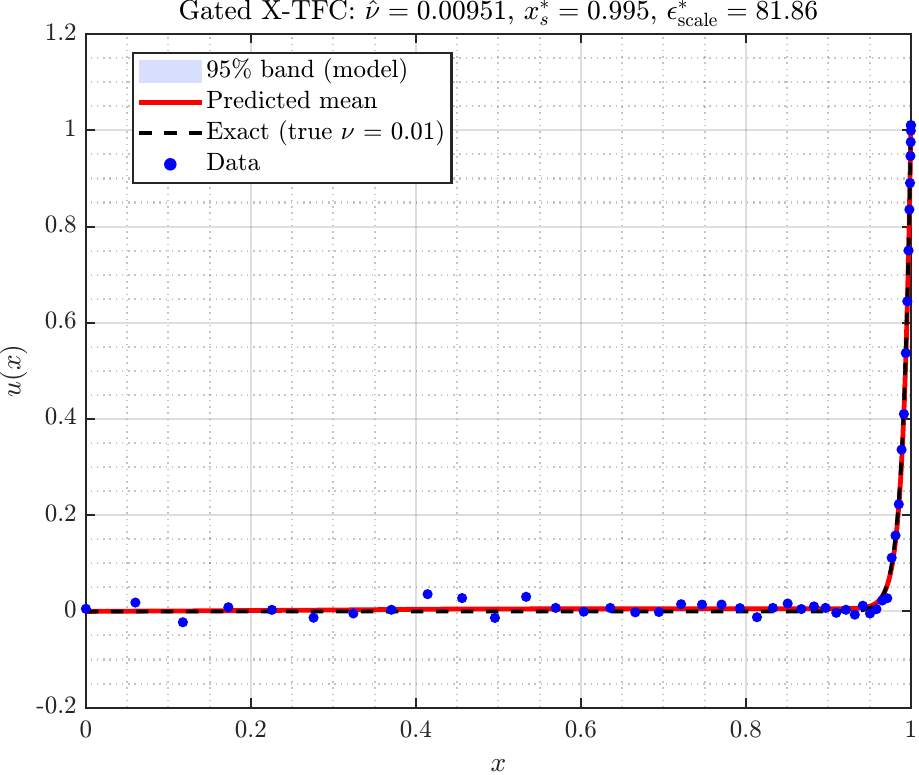}
		\caption{$\nu_{\text{true}}=10^{-2}$: the learned split drifts toward the outflow and concentrates resolution near $x=1$, yielding an accurate, stable fit.}
	\end{subfigure}
	
	\medskip
	
	\begin{subfigure}{0.48\textwidth}
		\includegraphics[width=\linewidth]{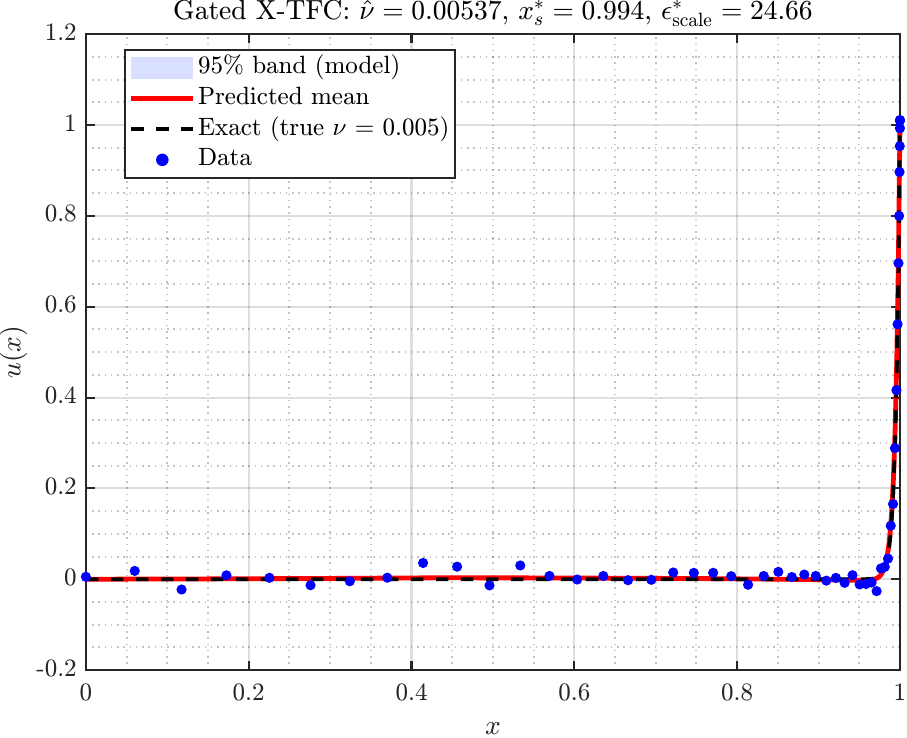}
		\caption{$\nu_{\text{true}}=5\times 10^{-3}$: a thinner layer induces a larger right shift in $x_s^\ast$ and a steeper gate; uncertainty remains small.}
	\end{subfigure}\hfill
	\begin{subfigure}{0.48\textwidth}
		\includegraphics[width=\linewidth]{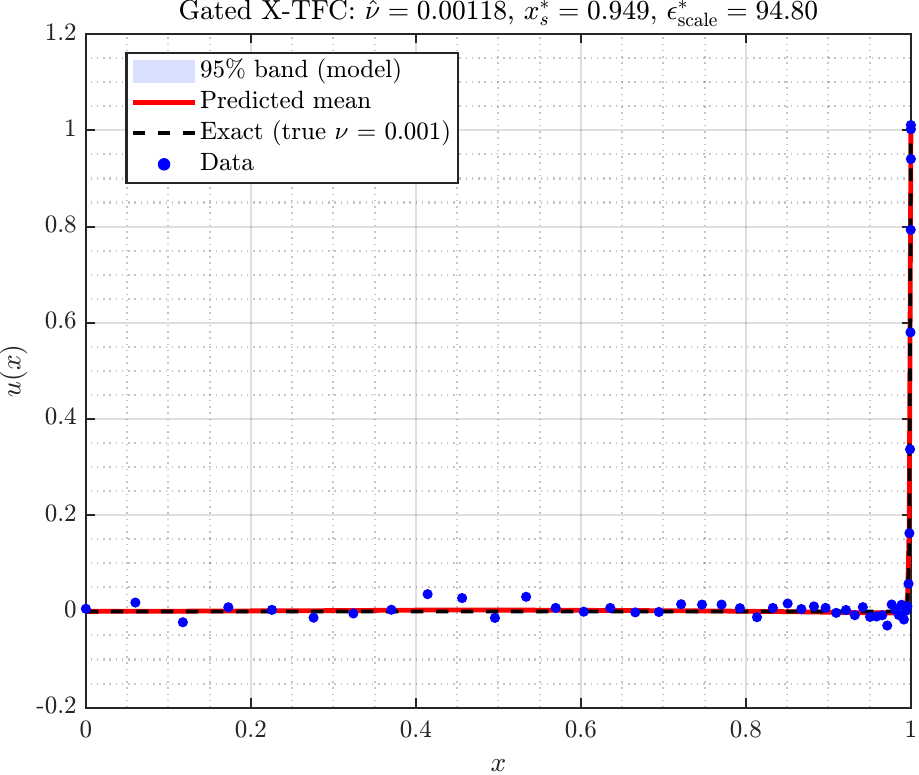}
		\caption{$\nu_{\text{true}}=10^{-3}$: extreme layer resolved without oscillations; panel title reports $\hat{\nu}$, $x_s^\ast$, and $\varepsilon_{\text{scale}}^\ast$.}
	\end{subfigure}
	
	\caption{Gated X--TFC inverse solutions across decreasing diffusion. Red: posterior mean; shaded: $95\%$ model band; blue: data; black dashed: exact (for reference only).}
	\label{fig:two-by-two}
\end{figure}

\begin{figure}[htbp]
	\centering
	\includegraphics[width=0.95\linewidth]{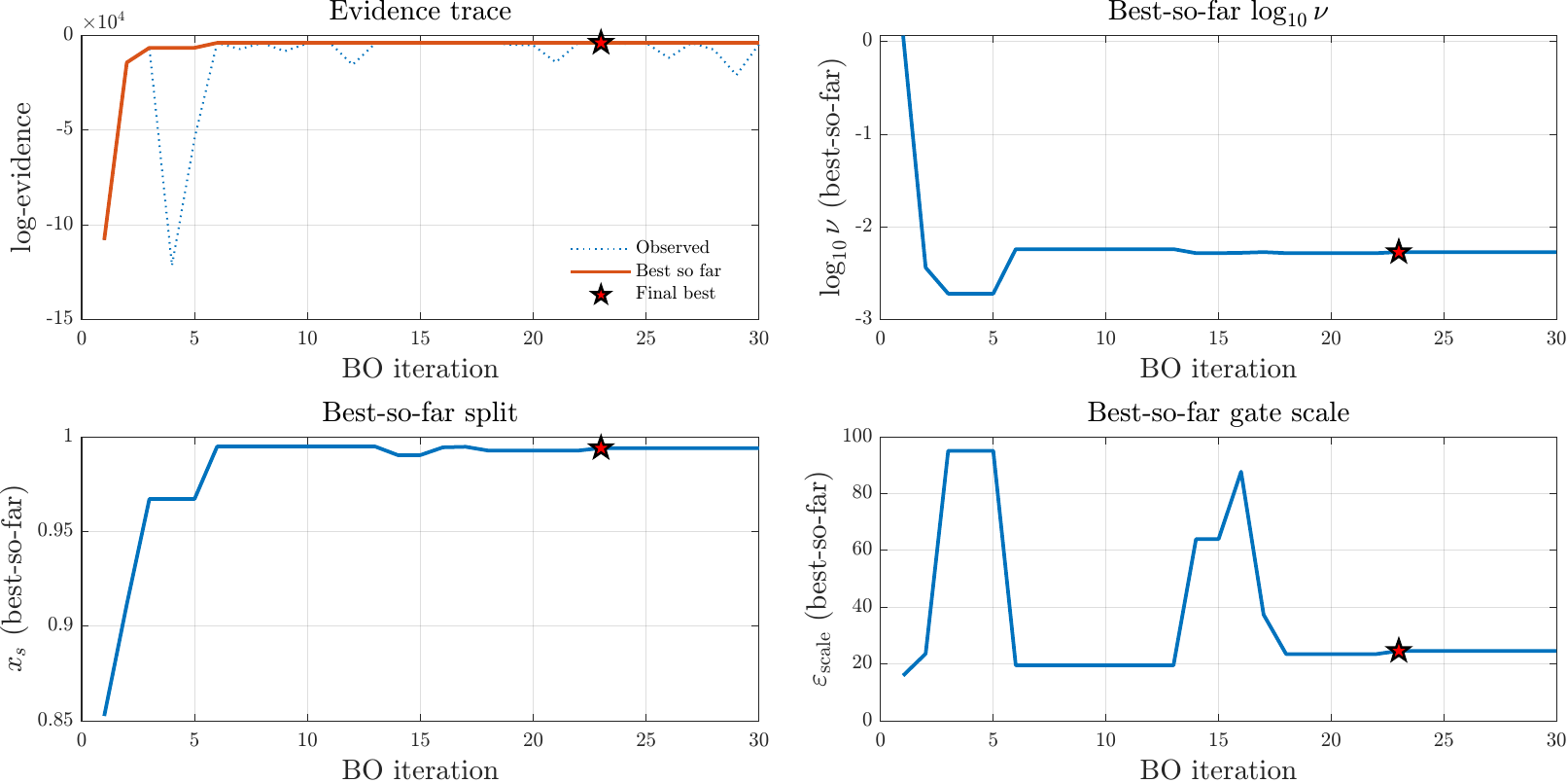}
	\caption{Bayesian optimization (BO) trace for the inverse problem ($\nu_{\text{true}}=5\times 10^{-3}$, $\nu_{\text{pred}}=5.37\times 10^{-3}$). 
		Top-left: log-evidence at each evaluation (dotted) and its best-so-far envelope (solid); the star marks the final maximum-evidence model. 
		Top-right/bottom: best-so-far hyperparameters aligned with the envelope—$\log_{10}\nu$, split $x_s$, and gate scale $\varepsilon_{\mathrm{scale}}$. 
		BO maximizes evidence; the starred parameters are used for the reported reconstruction.}
	\label{fig:INV_BayesOpt}
\end{figure}

We next consider the inverse problem: recover the governing parameter from noisy pointwise observations while jointly adapting the internal soft split and its transition scale. The inference is posed in a linear--Gaussian (whitened) setting, which allows us to evaluate the marginal likelihood (evidence) exactly and use it as a single, well-conditioned scalar objective for global search.

\paragraph{Experimental Setup}
For each test, we synthesize $N_{\text{data}}=50$ observations from the exact solution with additive Gaussian noise of standard deviation $10^{-2}$, sampled on a right-clustered grid (exponent $p=3$) to mimic boundary-layer sensing near $x=1$. The constrained part of the trial satisfies the Dirichlet data exactly, while the free coefficients are estimated by residual least squares on two softly decomposed blocks:
\begin{itemize}
	\item \textit{Collocation and basis:} $N_c=400$ collocation points and $N_\star=300$ RBF centers \emph{per} block (two blocks total). Gaussian widths follow $\sigma_L=k\,d_L$ and $\sigma_R=k\,d_R$ with $k=1.5$ and block spacings $d_L,d_R$ induced by the split. The blend uses a logistic gate with transition width $\varepsilon_b=\max\{\varepsilon_{\text{geo}},\,\varepsilon_{\text{phy}}\}$, where $\varepsilon_{\text{geo}}=\varepsilon_{\text{scale}}\min(d_L,d_R)$ and $\varepsilon_{\text{phy}}=5\,\nu$.
	\item \textit{Whitening:} data precision $\beta_{\text{data}}=1/\sigma^2$ with $\sigma=10^{-2}$ and PDE precision $\beta_{\text{pde}}=10^{2}$; both blocks are stacked into one weighted linear system.
\end{itemize}

\paragraph{Bayesian Optimization Objective}
We optimize over
\[
\log_{10}\nu \in [-3,1],\qquad x_s \in [0.85,0.995],\qquad \varepsilon_{\text{scale}} \in [10,100],
\]
using expected-improvement-plus for $30$ evaluations. For each triplet, we auto-tune the ridge parameter $\eta\!\in\![10^{-12},10^{-2}]$ by type-II maximum likelihood (with jittered Cholesky safeguards), then compute the exact log-evidence $\log p(y\mid \nu,x_s,\varepsilon_{\text{scale}},\eta)$. BO minimizes the negative log-evidence; thus the \emph{maximum}-evidence model is the final incumbent.

\begin{table}[htbp]
	\centering
	\caption{Inverse recovery of $\nu$: absolute percentage error $\big(100\,|\hat{\nu}-\nu_{\text{true}}|/\nu_{\text{true}}\big)$ and wall--clock time.}
	\label{tab:inv_nu_percent_error}
	\begin{tabular}{cccc}
		\toprule
		$\nu_{\text{true}}$ & $\hat{\nu}$ & \% error & Time (s) \\
		\midrule
		0.1   & 0.0897  & 10.3\% & 12.0 \\
		0.01  & 0.00951 &  4.9\% & 12.0 \\
		0.005 & 0.00537 &  7.4\% & 11.5 \\
		0.001 & 0.00117 & 17.0\% & 14.0 \\
		\bottomrule
	\end{tabular}
\end{table}

\paragraph{Results and Analysis}
Figure~\ref{fig:two-by-two} summarizes four cases from mild to extreme boundary layers, $\nu_{\text{true}}\in\{10^{-1},\,10^{-2},\,5\times 10^{-3},\,10^{-3}\}$. In all panels, the red curve is the posterior mean, the shaded region the $95\%$ model band, blue dots the noisy data, and the black dashed curve the exact solution (for reference only). As diffusion decreases, the learned split migrates toward $x=1$ and the gate steepens, concentrating width where the layer lives. Uncertainty remains tight near measurements and expands modestly only at the layer edge. Even for $\nu_{\text{true}}=10^{-3}$, the reconstruction is stable and free of oscillations; panel titles (last column) report the recovered parameter and gate settings. Quantitatively, Table~\ref{tab:inv_nu_percent_error} reports absolute percentage error and wall–clock time across the same four cases: errors are $\approx 10.3\%$ ($10^{-1}$), $4.9\%$ ($10^{-2}$), $7.4\%$ ($5\times 10^{-3}$), and $17.0\%$ ($10^{-3}$), with run times between $11.5$ and $14$ seconds on our hardware.

\paragraph{BO Dynamics}
The optimization trace for $\nu_{\text{true}}=5\times 10^{-3}$ (Fig.~\ref{fig:INV_BayesOpt}) illustrates the process. The top-left panel shows the observed log-evidence per evaluation (dotted) and its best-so-far envelope (solid); the star marks the final maximum-evidence model with $\nu_{\text{pred}}=5.37\times 10^{-3}$. The remaining panels align the best-so-far hyperparameters with this envelope, revealing a coherent progression toward the incumbent: $\log_{10}\nu$, the split location, and the gate scale stabilize jointly as evidence improves.

Evidence maximization supplies a single, principled scalar target that automatically balances data fit and model complexity while steering the operator and gate together. Coupled with the soft split, it yields accurate recovery and calibrated uncertainty across all regimes tested, without any interface penalties or multi-objective tuning.

\subsection {Operator-Conditioned Meta Learning}
\begin{figure}[htbp]
	\centering
	\begin{subfigure}[b]{0.5\textwidth}
		\centering
		\includegraphics[width=\linewidth]{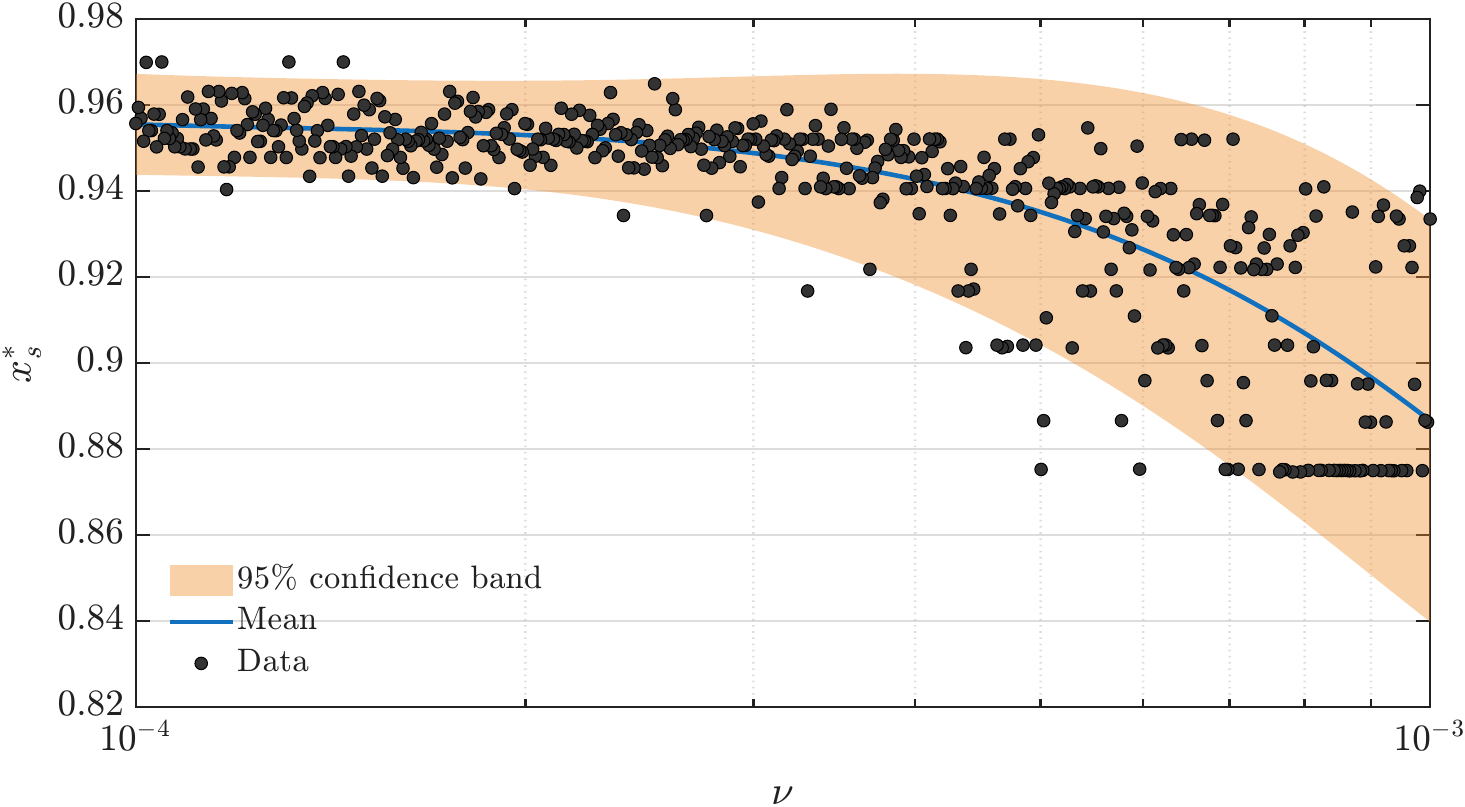}
		\caption{Operator–conditioned split location. Mean and 95\% band of $x_s^{\ast}$ vs.\ $\nu$ (log-$x$); dots are training optima. The band furnishes the search window $[x_s^{\mathrm{L}},x_s^{\mathrm{U}}]$.}
		\label{fig:fig1}
	\end{subfigure}
	\hfill
	\begin{subfigure}[b]{0.49\textwidth}
		\centering
		\includegraphics[width=\linewidth]{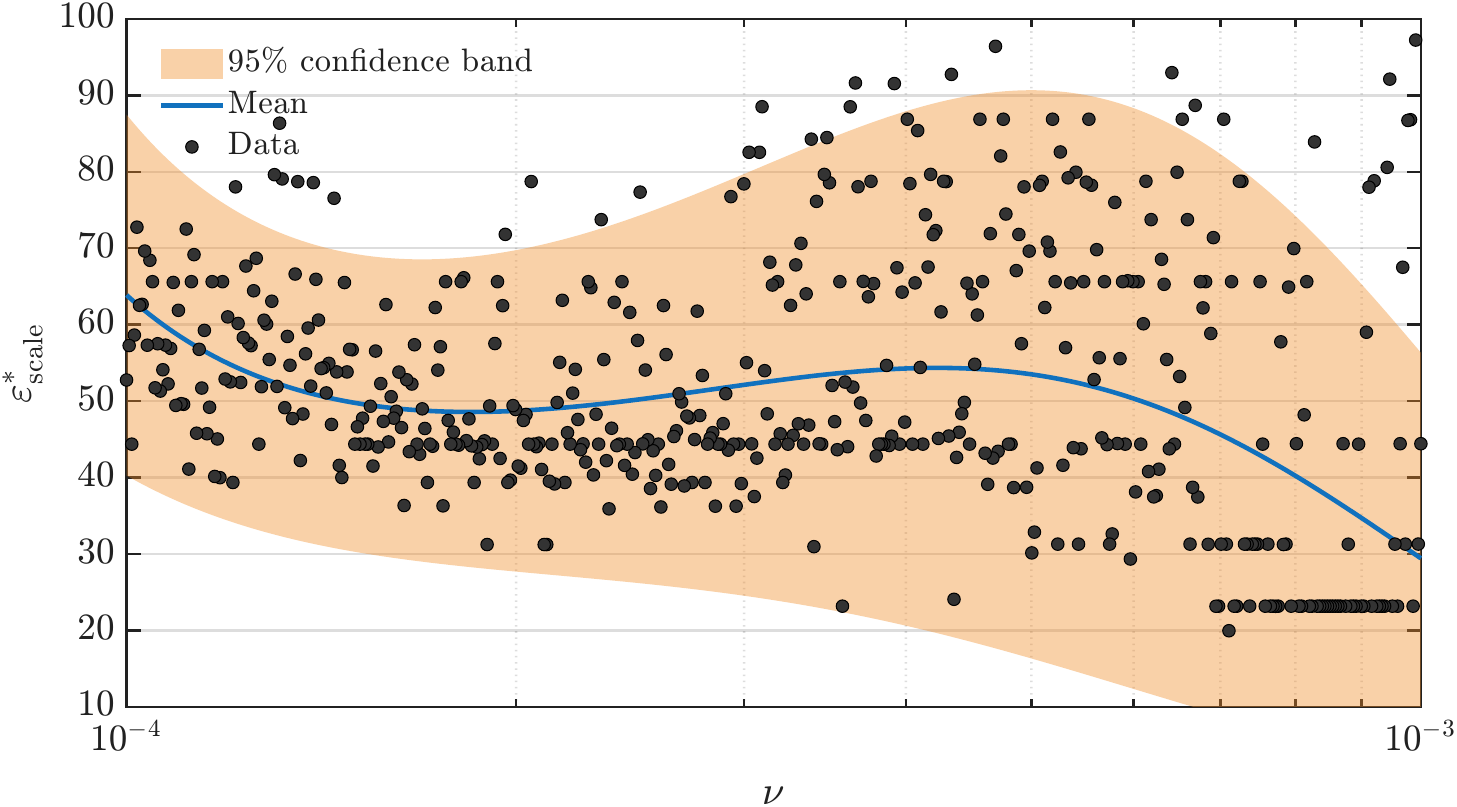}
		\caption{Operator–conditioned gate scale. Mean and 95\% band of $\varepsilon_{\mathrm{scale}}^{\ast}$ vs.\ $\nu$ (log-$x$); dots are training optima. The band yields $[\varepsilon^{\mathrm{L}},\varepsilon^{\mathrm{U}}]$.}
		\label{fig:fig2}
	\end{subfigure}
	\caption{Meta–learning maps from diffusivity to gate settings for Gated X–TFC. The uncertainty–aware predictions provide warm starts and tight, problem–specific bounds that accelerate the forward and inverse solvers.}
	\label{fig:combined}
\end{figure}
\begin{figure}[htbp]
	\centering
	\begin{subfigure}{0.65\textwidth}
		\centering
		\includegraphics[width=\linewidth]{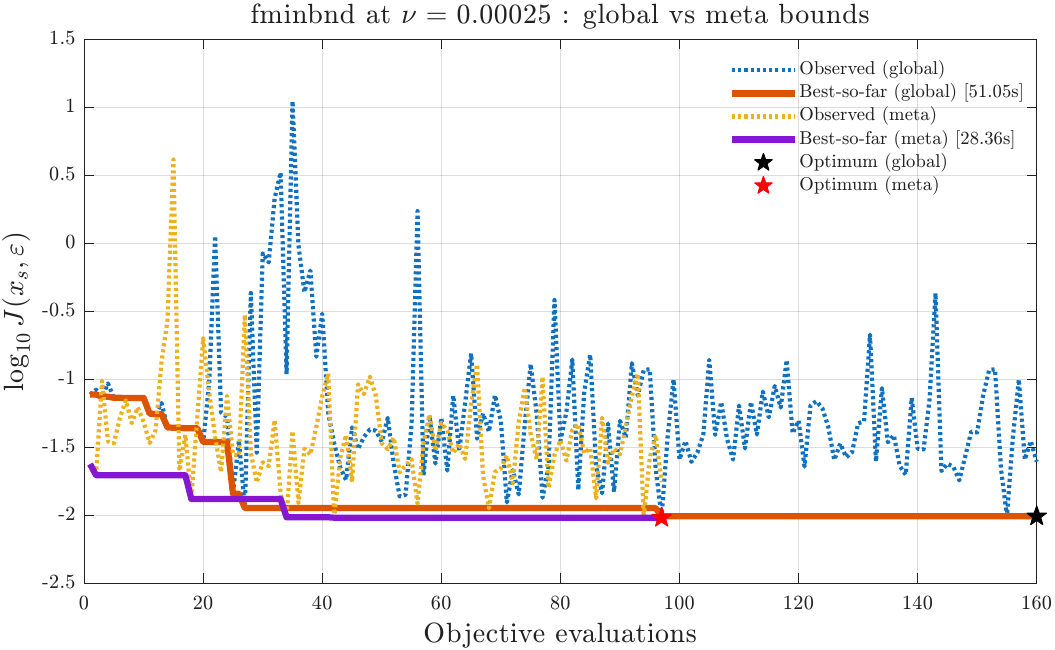}
	\end{subfigure}
	
	\begin{subfigure}{0.65\textwidth}
		\centering
		\includegraphics[width=\linewidth]{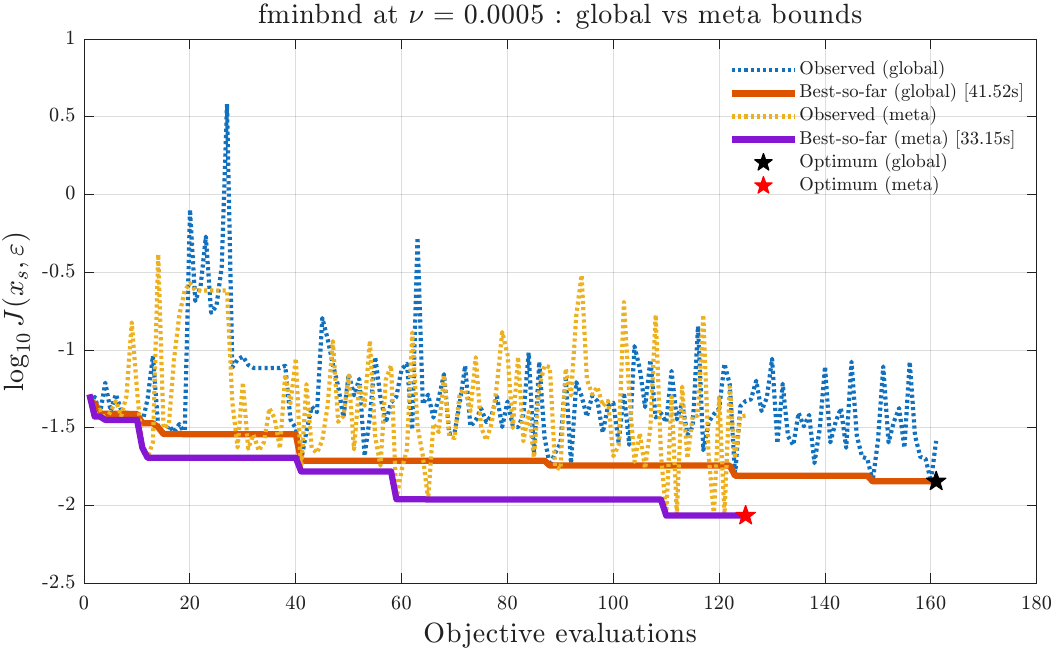}
	\end{subfigure}
	
	\begin{subfigure}{0.65\textwidth}
		\centering
		\includegraphics[width=\linewidth]{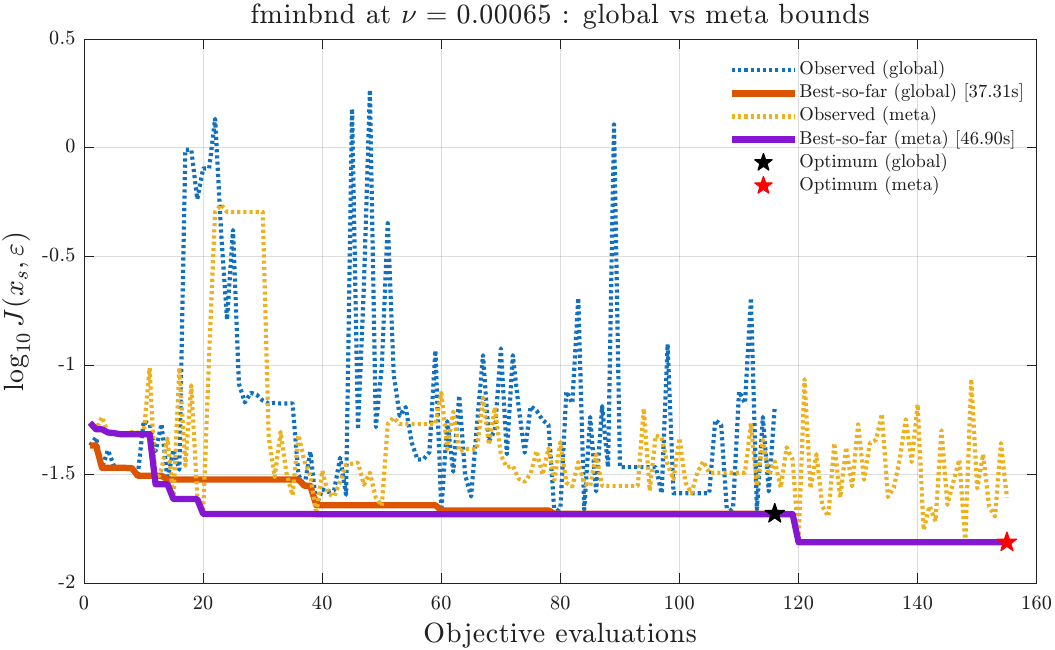}
	\end{subfigure}
	
	\caption{
		Performance evaluation of nested \texttt{fminbnd} optimization at $\nu \approx 10^{-4}$, comparing global and meta-bounded search strategies for parameters $(x_s,\varepsilon_{\mathrm{scale}})$. The log-scale objective function ($J$) is plotted against the number of evaluations (dotted lines show observed values, solid lines show the best-so-far, and pentagrams mark the final optima). Legend values indicate wall-clock time per run. The meta-bounded method constrains the search space, leading to faster convergence and superior solution accuracy compared to the global approach.
	}
	\label{fig:meta_low_nu_comparison}
	
\end{figure}

\paragraph{Training Dataset Generation}
We generate $500$ supervised samples by sweeping $\nu$ on a logarithmic grid and, for each value, running the same forward Gated X--TFC solver described in the previous section to obtain the bilevel optima. The global search bounds are $x_s \in [0.80,\,0.99]$ and $\varepsilon_{\text{scale}} \in [10,\,100]$. The resulting triples $\big(\nu,\,x_s^\ast,\,\varepsilon_{\text{scale}}^\ast\big)$ form the dataset used for meta-learning.

\paragraph{Heteroskedastic Meta–Regression}
We learn uncertainty–aware maps from the operator parameter to the gating hyperparameters using a cubic polynomial weighted ridge model on the remapped space ($x=\log_{10}\nu$, logit/log outputs). The fit is \emph{heteroskedastic}: we estimate a location–dependent noise level by Gaussian–smoothing squared residuals and iterate an iteratively reweighted least squares (IRLS) loop, selecting the ridge strength at each pass by weighted \emph{generalized cross–validation} (GCV). Predictive bands combine model variance with the local noise and are then back–transformed to physical units via the delta method (see \ref{app:hetero-reg} for full details). The resulting $95\%$ bands in Figs.~\ref{fig:fig1}–\ref{fig:fig2} tighten markedly near $\nu\approx10^{-4}$, enabling much narrower, problem–aware search boxes than the global ranges used to generate the dataset. For a related application of this physics-aware tuning bands in physics-informed image segmentation, we refer readers to \cite{Dwivedi2024}.

\paragraph{Speedups from Meta–Bounded Search}
We compare two search strategies for tuning the interface and gate scale, \((x_s,\varepsilon_{\mathrm{scale}})\), at very small diffusivities (\(\nu \approx 10^{-4}\)): (A) a \emph{global} box \([0.80,\,0.999]\times[10,\,100]\) and (B) a \emph{meta–bounded} box given by the \(95\%\) predictive intervals from our heteroskedastic meta–regressor (\ref{app:hetero-reg}), clipped to the same global box and softly padded when the band collapses (by \(5\times10^{-3}\) in \(x_s\) or \(0.5\) in \(\varepsilon_{\mathrm{scale}}\)). Both strategies minimize the \emph{same} objective—the strong-form residual energy \(J\) evaluated on an independent validation grid (800 points); the forward residual uses the layout from the forward section (\(N_c=N_\star=1000\) per block, \(k=1.5\), and the physics floor \(\varepsilon_{\mathrm{phy}}=5\nu\)). 

Figure~\ref{fig:meta_low_nu_comparison} plots \(\log_{10} J\) versus total objective evaluations for three representative \(\nu\) values near \(10^{-4}\). Dotted curves show the observed values at each evaluation; solid curves show the best-so-far envelopes; pentagrams mark the final optima. Legends report wall-clock times from the runs. Across all cases, the meta–bounded traces drop rapidly and stabilize earlier than the global traces. Constraining the search to operator-conditioned bands prunes implausible regions of \((x_s,\varepsilon_{\mathrm{scale}})\), yielding fewer evaluations, shorter wall time, and equal or lower terminal \(J\) than the global search—i.e., simultaneous speed and accuracy gains without changing the optimizer or the objective. 

\subsection{Extension to Multi-Domain and 2D Problems}
\label{sec:Extensions}
In this section, we present two applications of Gated X--TFC: (a) the twin boundary layer problem with multiple domain splits, and (b) the two-dimensional Poisson equation with a sharp source term.  
\begin{enumerate}
	\item \textit{Twin Boundary Layer Problem.} The governing equation, boundary conditions, and exact solution are given by:  
	\begin{itemize}
		\item PDE: $2(2x-1)u_x - \nu u_{xx} + 4u = 0, x\in[0,1]$
		\item BC: $u(0) = 1$, $u(1) = 1$
		\item Exact solution: $u(x) = e^{-2x(1-x)/\nu}$
	\end{itemize}
	This problem exhibits two boundary layers located at $x=0$ and $x=1$. As $\nu$ decreases, the boundary layers become increasingly sharp. 
	
	The problem formulation remains the same as before, except that two split locations are employed instead of one, with 1200 RBFs per block rather than 1000, and Bayesian optimization is used in place of \texttt{fminbnd} to improve computational speed.
	
	Figure~\ref{fig:twin_bl_exact_vs_xtfc} shows the Gated X--TFC solution compared with the exact result at a low value of $\nu = 10^{-4}$. The PDE residuals and absolute errors are presented in Figure~\ref{fig:twin_bl_residuals}, while the RBF width distribution is illustrated in Figure~\ref{fig:twin_bl_width_profile}. Total time taken is close to 50 seconds.
	\begin{figure}[htbp]
		\centering
		\includegraphics[width=0.8\linewidth]{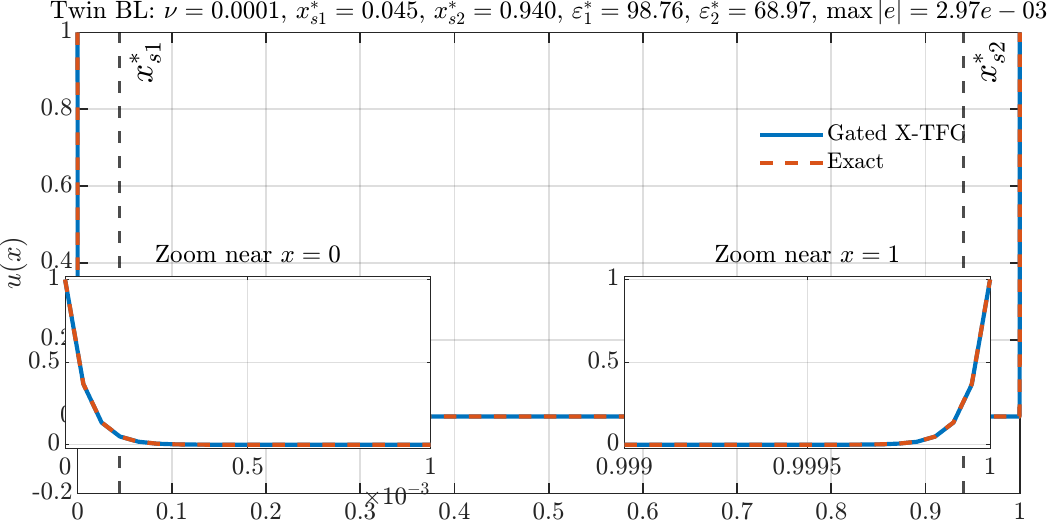} 
		\caption{Gated X--TFC vs. exact solution for the twin boundary layer case at $\nu = 10^{-4}$.}
		\label{fig:twin_bl_exact_vs_xtfc}
	\end{figure}
	
	\begin{figure}[htbp]
		\centering
		\includegraphics[width=0.6\linewidth]{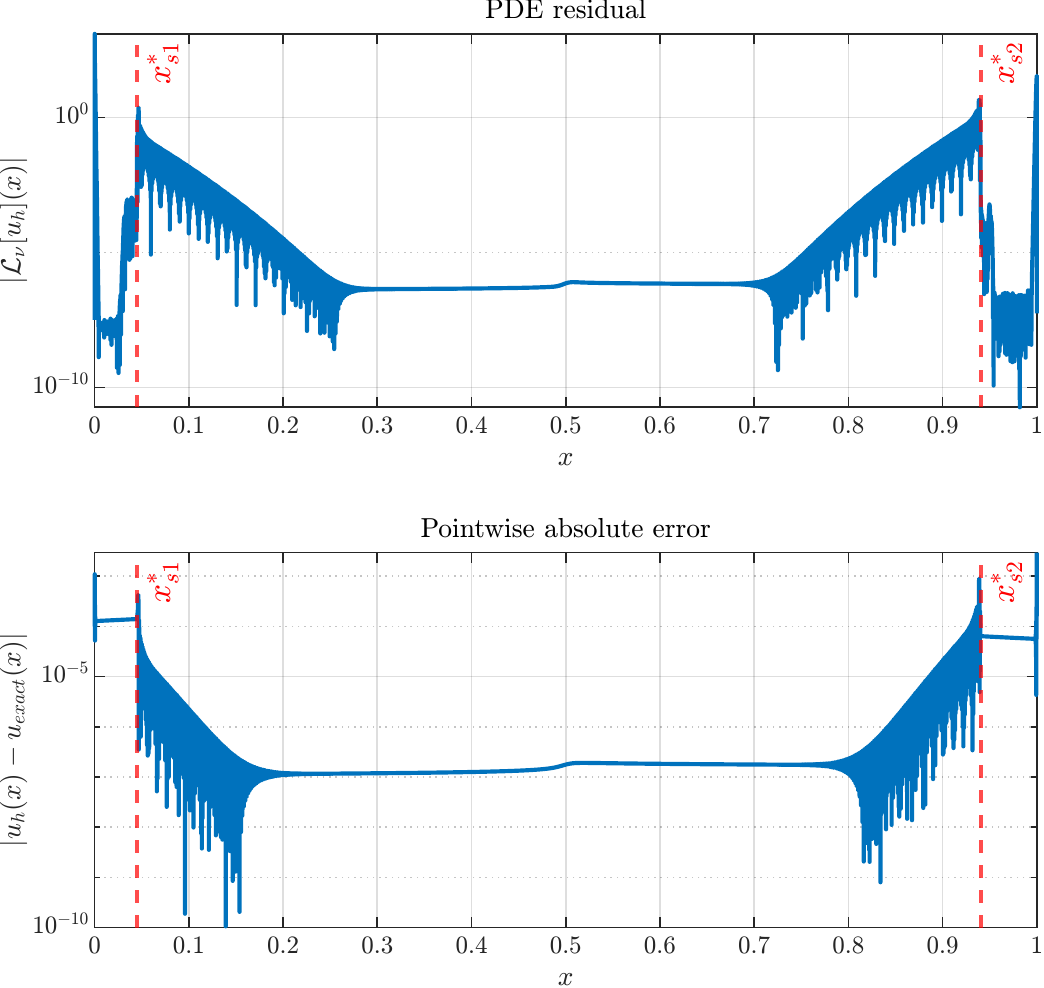} 
		\caption{Residual and absolute error, \(\nu=10^{-4}\).}
		\label{fig:twin_bl_residuals}
	\end{figure}
	
	\begin{figure}[htbp]
		\centering
		\includegraphics[width=0.6\linewidth]{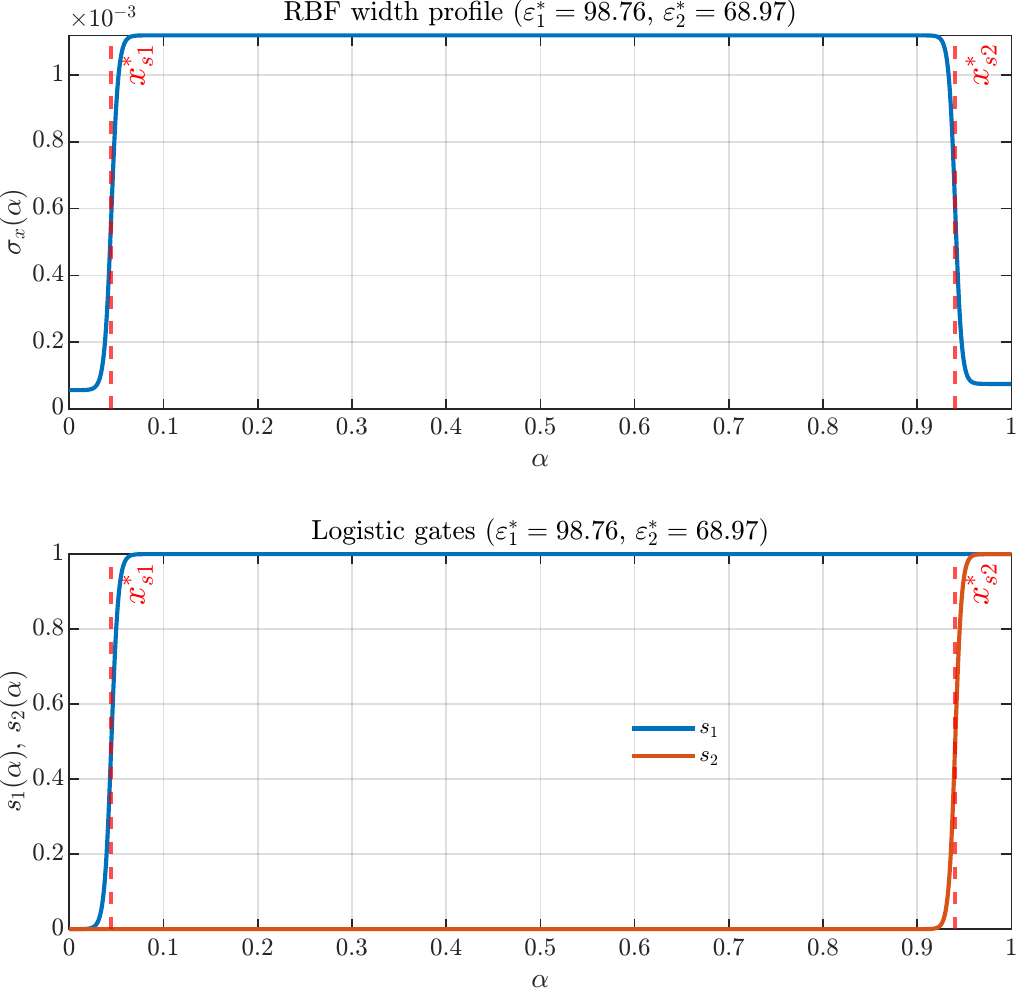} 
		\caption{Width and gate profiles, \(\nu=10^{-4}\).}
		\label{fig:twin_bl_width_profile}
	\end{figure}
	
	\item \textit{2D Poisson with Sharp Source Term.} We solve the Poisson equation over the unit square domain \(\Omega = [0,1]^2\) with homogeneous Dirichlet boundary conditions:
	\begin{equation}
		\begin{aligned}
			\nabla^2 u(x,y) &= S(x,y), \quad && (x, y) \in \Omega, \\
			u(x,y) &= 0, \quad && (x, y) \in \partial\Omega,
		\end{aligned}
		\label{eq:poisson_problem}
	\end{equation}
	
	where the source term \(S(x,y)\) is given by
	\begin{equation}
		S(x,y) = \frac{1}{2\pi\nu^2} \exp\left(-\frac{(x - 0.5)^2 + (y - 0.5)^2}{2\nu^2}\right).
		\label{eq:source_term}
	\end{equation}
	The Gaussian source is localized at the center of the domain, and for small values of $\nu$, it induces sharp gradients in the solution.
	\begin{figure}[htbp]
		\centering
		\includegraphics[width=0.8\linewidth]{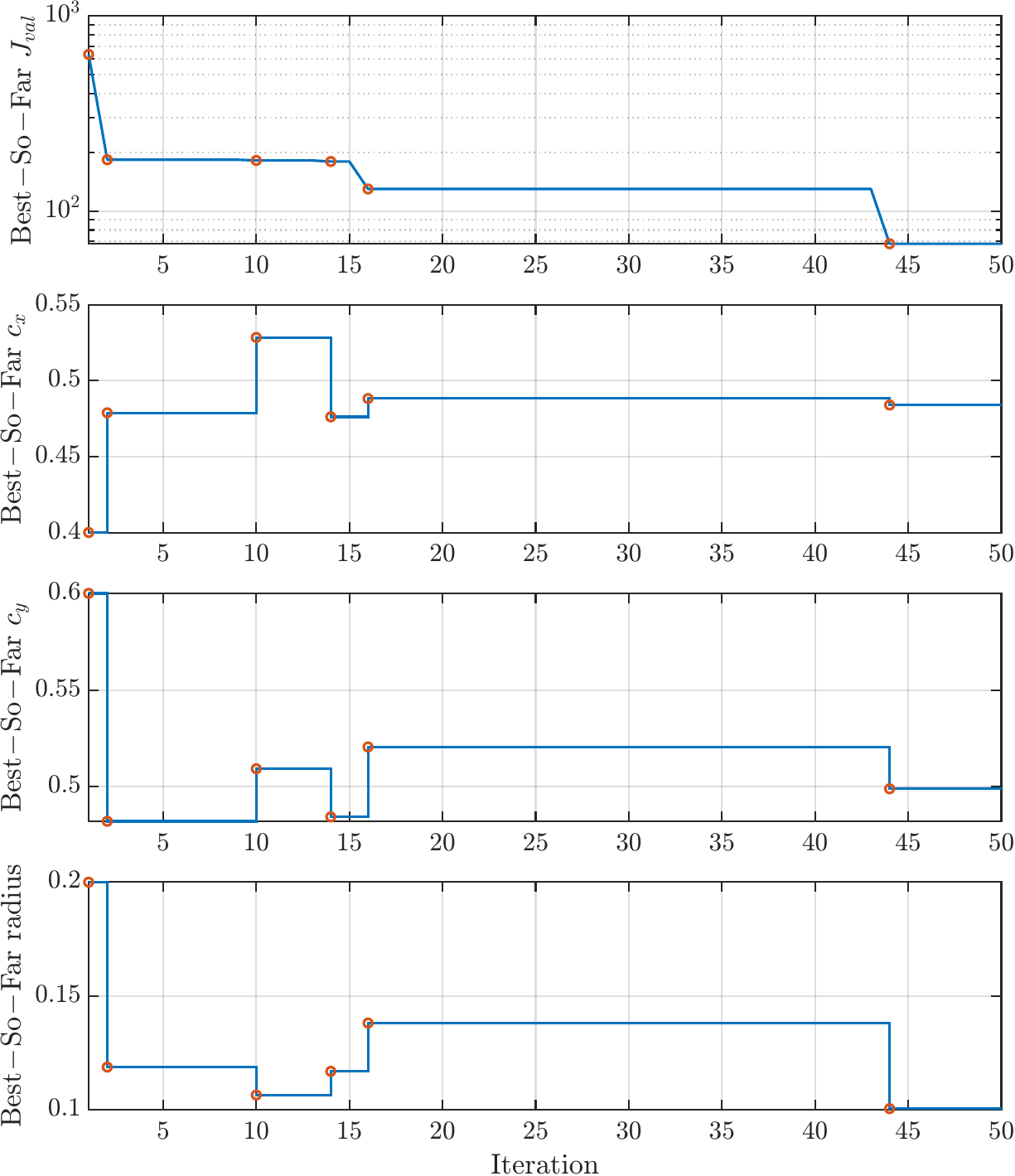} 
		\caption{Training history of hyperparameters for the 2D Poisson equation case at \(\nu=10^{-2}\).}
		\label{fig:2D_train_history}
	\end{figure} 
	\begin{figure}[htbp]
		\centering
		\includegraphics[width=0.99\linewidth]{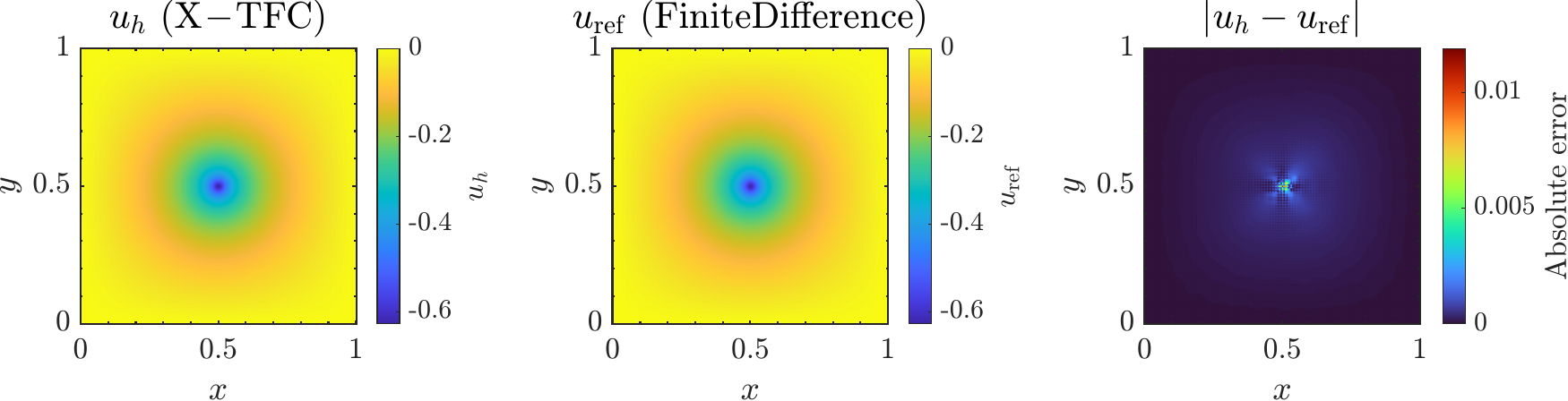} 
		\caption{Gated X--TFC vs. finite difference solution for the 2D Poisson equation case at \(\nu=10^{-2}\).}
		\label{fig:2D_exact_vs_pielm}
	\end{figure}
	
	\begin{figure}[htbp]
		\centering
		\begin{subfigure}{0.48\textwidth}
			\centering
			\includegraphics[width=\linewidth]{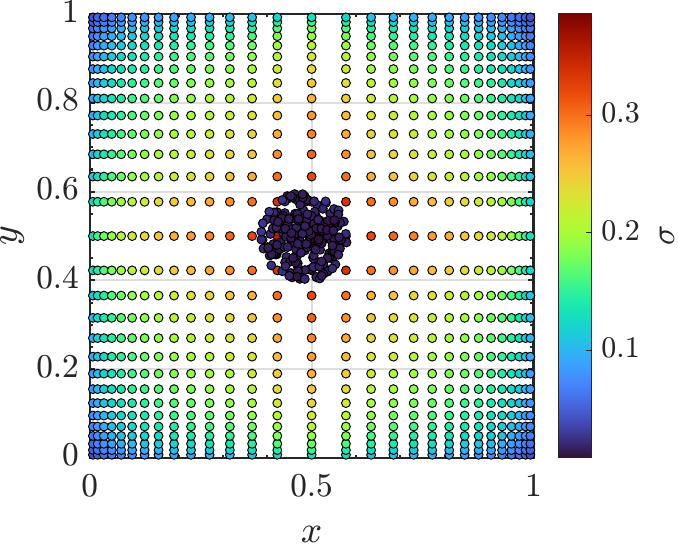}
			\caption{RBF width profile}
		\end{subfigure}%
		\hfill
		\begin{subfigure}{0.48\textwidth}
			\centering
			\includegraphics[width=\linewidth]{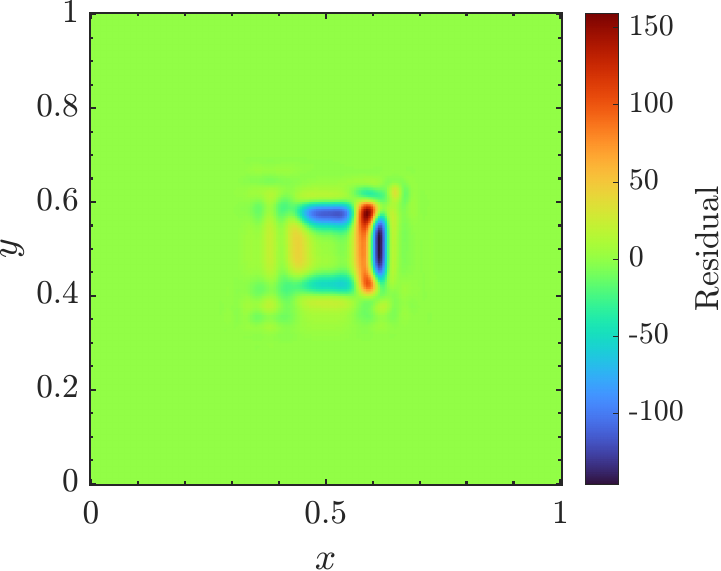}
			\caption{Point-wise PDE residual}
		\end{subfigure}
		\caption{RBF width profile and point-wise PDE residual at \(\nu=10^{-2}\).}
		\label{fig:2D_RBF_profile_Residual}
	\end{figure}

	\paragraph{Trial Function}
	To solve this equation, we construct a trial solution \citep{Lagaris} that exactly satisfies the boundary conditions:
	\begin{equation}
		u_h(x,y) = \sum_{i=1}^{N_s} c_i \, G(x,y) \, \phi_i(x,y),
		\label{eq:xtfc-hyp}
	\end{equation}
	where the coefficients \(c_i\) are to be determined and $G(x,y)=x(1-x)y(1-y)$.

	\paragraph{RBF Centers and Width Distribution}
	We build a two–scale mixture of isotropic Gaussian RBFs
	\[
	\phi_i(x,y)=\exp\!\big(-m_i^2\!\big[(x-x_i)^2+(y-y_i)^2\big]\big),\qquad 
	m_i=\frac{1}{\sqrt{2}\,\sigma_i},
	\]
	by combining a \emph{global} Chebyshev-like grid with a \emph{local} disk enrichment around a tunable center.
	
	\begin{itemize}
		\item \textit{Global coverage.} Along each axis we use a Chebyshev–linear blend
		\(
		t = (1-\alpha)\,t_{\text{lin}} + \alpha\,t_{\text{cheb}}
		\)
		to form a tensor grid \(\mathcal{C}_{\text{global}}\subset[0,1]^2\). The global grid is $31 \times 31.$
		A per-node width is set from the local grid spacing via the geometric mean,
		\[
		\sigma_{\text{global}}(x,y) \;=\; c_\sigma\,\sqrt{\Delta x\,\Delta y}, c_\sigma=5.
		\]
		
        \item \textit{Local enrichment.} We add \(N_{\text{in}}(=200)\) points uniformly \emph{by area} in a disk
        centered at \(w=(c_x,c_y)\) with radius \(r\). We sample \(r' = r\sqrt{U}\) and \(\theta=2\pi T\), where $U$ is  uniform random variable on [0,1] used to place points uniformly by area inside a disk of radius r, and similarly $T=2 \pi\times U$. Next, we set \((x,y)=(c_x+r'\cos\theta,\;c_y+r'\sin\theta)\), clipped to \([0,1]^2\).
        We denote this set by \(\mathcal{C}_{\text{local}}\).

        \item \textit{Local widths from kNN with global cap.} For each \((x_i,y_i)\in\mathcal{C}_{\text{local}}\),
        let \(d_k(i)\) be the distance to its \(k\)-th nearest neighbor (here \(k=6\)).
         We define a kNN-based width with mild clipping
        \[
         \sigma_i^{\text{knn}} \;=\; \operatorname{clip}\!\big(0.9d_k(i),\;[\,0.25\,h_{\text{ref}},\;4\,h_{\text{ref}}\,]\big),
         \qquad h_{\text{ref}}=\operatorname{median}_i d_k(i),
         \]
         and cap it by the global scale interpolated at that location,
         \[
           \sigma_i \;=\; \min\!\big(\sigma_i^{\text{knn}},\;\sigma_{\text{global}}(x_i,y_i)\big).
         \]
		
		\item \textit{Final mixture.} The centers and widths are
		\[
		\mathcal{C}=\mathcal{C}_{\text{global}}\cup\mathcal{C}_{\text{local}}, 
		\]
		yielding a basis with coarse global coverage plus focused resolution within the local disk, while keeping local widths consistent with the ambient global spacing. This method is slightly different from the previous gated approach in that it does not learn a logistic gate explicitly; instead, it combines the global and local distributions through superposition.
	\end{itemize}
	\paragraph{Optimization}
	In the previous examples, we noticed that as the gradient steepens, the logistic gate saturates and effectively becomes hard.  We therefore adopt a hard logistic gate from the outset and delegate tuning to an  outer Bayesian optimization loop that adjusts only the local disk parameters—
	its center \((c_x,c_y)\) and radius \(r\). The training history of these 
	hyperparameters (Figure~\ref{fig:2D_train_history}) shows the disk migrating from 
	\((c_x,c_y)=(0.4,\,0.6)\) toward the domain center while contracting from 
	\(r=0.2\) to \(r=0.1\), which enables accurate capture of the sharp jump. To avoid overfitting, the training and validation grids are kept independent like before. 
	
	\paragraph{Results}
	Figures~\ref{fig:2D_exact_vs_pielm} and~\ref{fig:2D_RBF_profile_Residual} present, respectively, a comparison of the gated X--TFC solution with the finite–difference reference, and the pointwise PDE residuals together with the optimal RBF width map. The gated X--TFC solution agrees closely with the finite–difference result, and—as in prior experiments—the RBF widths show a marked contrast between the global background grid and the locally enriched region. 
\end{enumerate}

\subsection{Limitations and Future Work}

\begin{enumerate}
	\item Like most physics-informed methods, both X--TFC and Gated X--TFC encounter difficulties when dealing with extremely sharp gradients. The gating mechanism provides partial relief by adaptively focusing resolution, but further refinement is necessary to ensure robustness across diverse singular perturbation regimes.
	
	\item The present study has mainly demonstrated proof-of-concept on problems partitioned into at most three softly defined subdomains. Extending the approach to more complex geometries and an arbitrary number of partitions in a scalable way remains an open challenge.
	
	\item Although the elimination of boundary and interface penalties simplifies optimization, the current formulation has not yet been generalized to irregular or arbitrary geometries. Addressing this limitation will require the development of more sophisticated constrained expressions.
	
	\item While both X--TFC and Gated X--TFC scale favorably with dimensionality and could, in principle, be extended to nonlinear problems via iterative least-squares schemes, the present work has focused exclusively on linear cases. A systematic investigation into nonlinear PDEs, including convergence guarantees and efficient solvers, represents an important direction for future research.
\end{enumerate}

\FloatBarrier

\section{Conclusion}
\label{sec:Conclusion}
We presented \emph{Gated X--TFC}, a simple and effective extension of X--TFC that tackles
singularly perturbed PDEs via a \emph{soft}, learned domain decomposition. By introducing
differentiable logistic gates, the method preserves X--TFC's exact boundary-condition
enforcement while eliminating interface penalties and multi–objective tradeoffs. The gates
modulate RBF widths across the domain, enabling crisp resolution of boundary layers with
substantial computational savings. On a 1D convection–diffusion benchmark, Gated X--TFC
achieves an order–of–magnitude lower error than standard X--TFC, uses \(80\%\) fewer
collocation points, and reduces training time by \(66\%\). Beyond deterministic forward
solves, an operator–conditioned meta–learner delivers uncertainty–aware warm starts and
tight search bounds for an outer Bayesian optimization loop, yielding a unified pipeline
for forward \emph{and} inverse problems. The approach scales naturally to multi–domain
and higher–dimensional settings, as demonstrated on a twin boundary–layer equation and
a 2D Poisson problem with a sharp Gaussian source.

In summary, the main highlights of this paper are as follows:
\begin{itemize}
	\item \textbf{Methodology.} Soft, gate–based decomposition within X--TFC removes backpropagation and interface penalties while keeping boundary conditions exact.
	\item \textbf{Efficiency.} Order–of–magnitude accuracy gains with \(80\%\) fewer
	collocation points and \(66\%\) shorter runtimes on a stiff 1D benchmark convection diffusion equation.
	\item \textbf{Unified Framework.} Unified framework for both forward and inverse problems.
	\item \textbf{Operator-Conditioned Meta Learning.} Uncertainty–aware warm starts and \emph{tight} search bounds for faster convergence over parameterized PDEs.
	\item \textbf{Scalability.} The construction extends cleanly to multiple-domains and to higher-dimensions.
\end{itemize}

Future work will target nonlinear PDEs, multi–gate/adaptive partition schedules,
theory for approximation and stability under gating, and tighter integration with
fast solvers and hardware acceleration.

\appendix
\section{Boundary–Layer Scale Analysis}
\label{app1}
For $0<\nu\ll1$, the outer solution satisfies $u'\approx0$ (constant), so the mismatch with $u(1)=1$ is resolved in a thin layer adjacent to the outflow boundary $x=1$ \citep{BALAJI2021121}. Let the layer thickness be $\delta\ll1$ and introduce the stretched coordinate
\[
\xi=\frac{1-x}{\delta}\,,\qquad 
u'(x)=-\frac{1}{\delta}u_\xi\,,\quad
u''(x)=\frac{1}{\delta^2}u_{\xi\xi}.
\]
Substituting into $u'-\nu u''=0$ gives
\[
-\frac{1}{\delta}u_\xi-\nu\,\frac{1}{\delta^2}u_{\xi\xi}=0
\;\;\Longleftrightarrow\;\;
u_\xi+\frac{\nu}{\delta}\,u_{\xi\xi}=0.
\]
Within the layer the convective and diffusive terms must be comparable, i.e., $u_\xi=O(1)$ and $(\nu/\delta)u_{\xi\xi}=O(1)$, which implies
\[
\frac{\nu}{\delta}=O(1)\quad\Rightarrow\quad \delta=O(\nu).
\]
Thus the boundary–layer thickness scales linearly with the viscosity: $\delta\sim\nu$.
\section{Heteroskedastic probabilistic meta–regression}
\label{app:hetero-reg}

We learn uncertainty–aware maps from the operator parameter to the gating hyperparameters using a lightweight, toolbox–free regression pipeline. Let the dataset be
\[
\mathcal{D} \;=\; \bigl\{(\nu_i,\,x^{\ast}_{s,i},\,\varepsilon^{\ast}_i)\bigr\}_{i=1}^{N}, 
\qquad N=500,
\]
constructed with the forward solver described in Section~\ref{sec:results-forward} under global search ranges
\(x_s\in[0.80,0.999]\) and \(\varepsilon_{\mathrm{scale}}\in[10,100]\).

\paragraph{Remapping to \( \mathcal{O}(1) \) space}
We regress on the log–scaled operator parameter
\[
x \;=\; \log_{10}\nu,
\]
and transform the targets to unbounded, numerically benign spaces:
\[
y_x \;=\; \operatorname{logit}\!\left(\frac{x_s^{\ast}-a_x}{\,b_x-a_x\,}\right), 
\qquad
y_{\varepsilon} \;=\; \log_{10}\varepsilon^{\ast}_{\mathrm{scale}},
\]
with anchors \(a_x\approx0.80\) and \(b_x\approx0.999\). These transforms preserve order, improve conditioning, and guarantee valid physical ranges after inverse mapping.

\paragraph{Weighted ridge model}
For each target \(y\in\{y_x,\,y_{\varepsilon}\}\) we fit a cubic polynomial ridge model \citep{montgomery2021introduction}
\begin{equation}
	\label{eq:app-model}
	y \;\approx\; \Phi(x)\,\beta,
	\qquad
	\Phi(x) \;=\; \bigl[\,1,\; x,\; x^2,\; x^3\,\bigr],
\end{equation}
by minimizing the weighted ridge objective
\begin{equation}
	\label{eq:app-ridge}
	\hat{\beta}_{\lambda}
	\;=\;
	\arg\min_{\beta}
	\;\bigl\|W^{1/2}\bigl(y-\Phi\beta\bigr)\bigr\|_2^2
	+\lambda\|\beta\|_2^2,
	\qquad
	W=\operatorname{diag}(w_1,\ldots,w_N).
\end{equation}
The closed form is 
\(
\hat{\beta}_{\lambda}
=
\bigl(\Phi^{\!\top}W\Phi+\lambda I\bigr)^{-1}\Phi^{\!\top}Wy.
\)

\paragraph{Heteroskedastic IRLS}
Because noise varies with \(x=\log_{10}\nu\), we use an iteratively reweighted least squares (IRLS) scheme \citep{davidian1987variance,goldberg1997regression,bishop2006pattern}:
\begin{enumerate}
	\item Initialize \(w_i\leftarrow 1\).
	\item Given \(W\), choose \(\lambda\) (below) and compute \(\hat{\beta}_{\lambda}\).
	\item Compute residuals \(r_i=y_i-\phi(x_i)^{\!\top}\hat{\beta}_{\lambda}\) and estimate local variance by Gaussian kernel smoothing,
	\begin{equation}
		\label{eq:app-smoother}
		\widehat{\sigma}^{2}(x)
		\;=\;
		\frac{\sum_{j=1}^N K_h(x-x_j)\,r_j^{\,2}}
		{\sum_{j=1}^N K_h(x-x_j)},
		\qquad 
		K_h(t)=\exp\!\bigl(-\tfrac{1}{2}(t/h)^2\bigr),
	\end{equation}
	with bandwidth \(h\) taken as a fixed fraction of the \(x\)-range.
	\item Update \(w_i \leftarrow 1/\max\{\widehat{\sigma}^{2}(x_i),\,\varepsilon\}\) (small floor \(\varepsilon>0\)) and repeat Steps 2–4 for a few passes (typically three).
\end{enumerate}

\paragraph{Ridge selection by weighted GCV}
At each IRLS pass we set \(\lambda\) by minimizing weighted generalized cross–validation (GCV \citep{golub1979generalized}),
\begin{equation}
	\label{eq:app-gcv}
	\mathrm{GCV}(\lambda)
	\;=\;
	\frac{\displaystyle \sum_{i=1}^N w_i\,\bigl(y_i-\phi(x_i)^{\!\top}\hat{\beta}_{\lambda}\bigr)^2}
	{\bigl(N-\operatorname{tr}S_{\lambda}\bigr)^2},
	\qquad
	S_{\lambda}
	\;=\;
	\Phi\bigl(\Phi^{\!\top}W\Phi+\lambda I\bigr)^{-1}\Phi^{\!\top}W,
\end{equation}
where \(S_{\lambda}\) is the weighted ridge hat matrix and \(\operatorname{tr}S_{\lambda}\) acts as the effective degrees of freedom.

\paragraph{Predictive mean and variance on the transformed scale}
For a query \(x_{\star}\) with feature vector \(\phi_{\star}=\Phi(x_{\star})^{\!\top}\),
\[
\mu_{\star} \;=\; \phi_{\star}^{\!\top}\hat{\beta}_{\hat{\lambda}},
\qquad
s_{\mathrm{model}}^{2}(x_{\star}) 
\;=\; \phi_{\star}^{\!\top}\,\widehat{\operatorname{Cov}}(\hat{\beta})\,\phi_{\star},
\qquad
s_{\mathrm{noise}}^{2}(x_{\star})
\;=\;\widehat{\sigma}^{2}(x_{\star}),
\]
with a sandwich-style coefficient covariance
\begin{equation}
	\label{eq:app-sandwich}
	\widehat{\operatorname{Cov}}(\hat{\beta})
	\;\approx\;
	\bigl(\Phi^{\!\top}W\Phi+\hat{\lambda}I\bigr)^{-1}
	\Phi^{\!\top}W\,\widehat{\Sigma}\,W\Phi
	\bigl(\Phi^{\!\top}W\Phi+\hat{\lambda}I\bigr)^{-1},
	\qquad
	\widehat{\Sigma}=\operatorname{diag}\!\bigl(\widehat{\sigma}^{2}(x_i)\bigr).
\end{equation}
The total predictive variance is
\(
s^{2}(x_{\star}) = s_{\mathrm{model}}^{2}(x_{\star}) + s_{\mathrm{noise}}^{2}(x_{\star})
\).
A two–sided \((1-\alpha)\) interval on the transformed scale is
\begin{equation}
	\label{eq:app-ci-transformed}
	\mu_{\star} \;\pm\; z_{1-\alpha/2}\, s(x_{\star}),
	\qquad
	z_{1-\alpha/2}=\Phi^{-1}(1-\alpha/2),
\end{equation}
with \(\Phi^{-1}\) the standard normal quantile function.

\paragraph{Back–transforms and delta–method bands}
We map predictions and intervals to physical units with monotone transforms.
\begin{align}
	\text{Split: } \quad
	\widehat{x}_s &= a_x + (b_x - a_x) \sigma(\mu_{\star}), \label{eq:app-back-xs} \\
	\sigma(t) &= \frac{1}{1 + e^{-t}}, \notag \\
	s_{x_s} &\approx (b_x - a_x) \sigma(\mu_{\star}) \bigl(1 - \sigma(\mu_{\star})\bigr) s(x_{\star}), \notag
\end{align}

\begin{align}
	\text{Scale: }\quad
	\widehat{\varepsilon}_{\mathrm{scale}}
	&= 10^{\,\mu_{\star}},
	&
	s_{\varepsilon}
	&\approx \ln(10)\,10^{\,\mu_{\star}}\,s(x_{\star}).
	\label{eq:app-back-eps}
\end{align}
We then report
\(
\widehat{x}_s \pm z_{1-\alpha/2}\, s_{x_s}
\)
clipped to \([a_x,b_x]\), and
\(
\widehat{\varepsilon}_{\mathrm{scale}} \pm z_{1-\alpha/2}\, s_{\varepsilon}
\)
clipped to broad positive bounds (e.g.\ \([10,100)\)).

\paragraph{Practical notes}
All quantities are remapped to \(\mathcal{O}(1)\) scales, which improves conditioning and numerics. The Gaussian smoother in~\eqref{eq:app-smoother} is robust and avoids toolbox dependencies; three IRLS passes suffice in practice. The resulting \(95\%\) bands tighten where data are informative (notably near \(\nu\approx 10^{-4}\)), enabling narrow, problem–aware search boxes for downstream solvers.

\section*{CRediT authorship contribution statement}
\textbf{Vikas Dwivedi:} Conceptualization, Methodology, Software, Writing - Original Draft, 
\textbf{Enrico Schiassi:} Methodology, Software, \textbf{Bruno Sixou} and \textbf{Monica Sigovan:} Supervision and Writing - Review \& Editing.

\section*{Acknowledgements}
This work was supported by the ANR (Agence Nationale de la Recherche), France, through the RAPIDFLOW project (Grant no. ANR-24-CE19-1349-01). 

\bibliographystyle{unsrt}  
\bibliography{references}

\end{document}